\documentclass[runningheads]{llncs}

\usepackage[review,year=2026,ID=10825]{eccv}
\usepackage{eccvabbrv}

\usepackage{graphicx}
\usepackage{booktabs}
\usepackage{booktabs}   % for \toprule, \midrule, \bottomrule, \cmidrule
\usepackage{multirow}   % for \multirow
\usepackage[table]{xcolor}
\usepackage[accsupp]{axessibility}  % Improves PDF readability for those with disabilities.

\usepackage{hyperref}

\usepackage{orcidlink}

\usepackage{amsmath}
\usepackage{graphicx}
\usetikzlibrary{
    arrows.meta,
    positioning,
    calc
}
\usepackage{graphicx}
\usepackage{subcaption}
\usepackage{amsmath, amssymb}

\usepackage{algorithm}
\usepackage{algorithmic}
\usepackage{pifont}
\definecolor{noirblue}{RGB}{70,130,180}

\AtBeginDocument{\nolinenumbers}

\tikzset{
    space/.style={
        draw=noirblue,
        very thick,
        ellipse,
        minimum width=2.6cm,
        minimum height=0.cm
    },
    point/.style={
        circle,
        fill=noirblue,
        inner sep=2.2pt
    },
    map/.style={
        -{Stealth[length=3mm]},
        ultra thick,
        draw=noirblue
    },
    correspondence/.style={
        -{Stealth[length=2.5mm]},
        dashed,
        thick,
        draw=gray
    },
    paneltitle/.style={
        font=\bfseries\small
    },
    sublabel/.style={
        font=\bfseries\tiny
    },
    img/.style={
        draw,
        ellipse,
        fill=black!8,
        minimum width=2.3cm,
        minimum height=1.7cm
    }
}

\begin{document}

\title{NOIR: Neural Operator Mapping for \\Implicit Representations} 

\titlerunning{NOIR}

\author{Sidaty El Hadramy\inst{1}\orcidlink{0009−0000−2917−0706} \and
Nazim Haouchine\inst{2}\orcidlink{0000-0002-1752-3479} \and
Michael Wehrli\inst{1}\orcidlink{0009−0005−8740−3295} \and  \\
Philippe C. Cattin\inst{1}\orcidlink{0000−0001−8785−2713}}

\authorrunning{El Hadramy et al.}

\institute{Department of Biomedical Engineering, University of Basel, Allschwil, Switzerland \and
Harvard Medical School, Brigham and Women's Hospital, Boston, United States}

\maketitle
\begin{figure}
    \centering
    \includegraphics[width=0.9\linewidth]{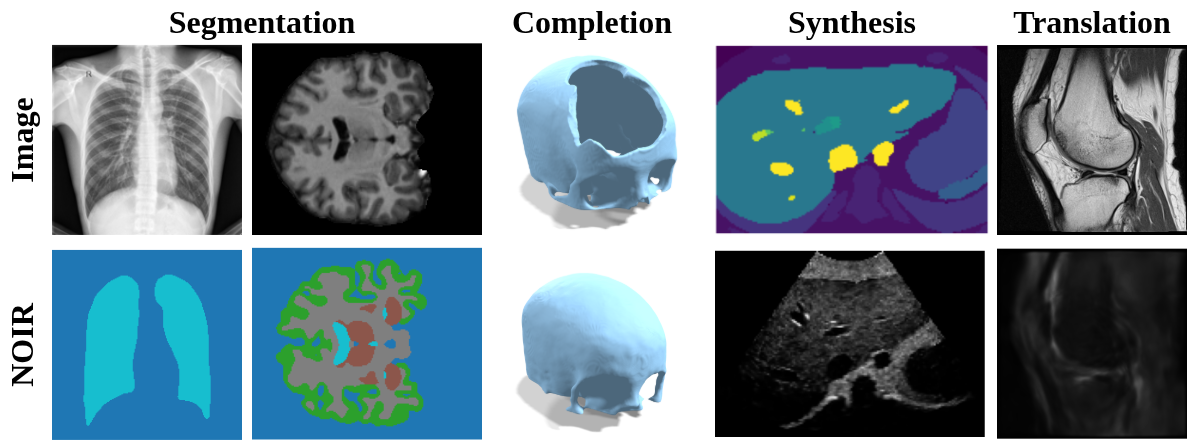}
    \caption{\textbf{Downstream tasks solved by NOIR}. By formulating each task as operator learning between function spaces, NOIR addresses segmentation, completion, synthesis, and translation across in 2D and 3D, with inherent robustness to resolution changes.}
    \label{fig:noir_overview}
\end{figure}

\begin{abstract}
 This paper presents NOIR, a framework that reframes core medical imaging tasks as operator learning between continuous function spaces, challenging the prevailing paradigm of discrete grid-based deep learning. Instead of operating on fixed pixel or voxel grids, NOIR embeds discrete medical signals into shared Implicit Neural Representations and learns a Neural Operator that maps between their latent modulations, enabling resolution-independent function-to-function transformations. We evaluate NOIR across multiple 2D and 3D downstream tasks, including segmentation, shape completion, image-to-image translation, and image synthesis, on several public datasets such as Shenzhen, OASIS-4, SkullBreak, fastMRI, as well as an in-house clinical dataset. It achieves competitive performance at native resolution while demonstrating strong robustness to unseen discretizations, and empirically satisfies key theoretical properties of neural operators. The project page is available here: \url{https://github.com/Sidaty1/NOIR-io}.
\end{abstract}

\section{Introduction}
\label{sec:intro}

\label{sec:intro}
Medical image computing is a specialized subfield of computer vision that aims to computationally interpret medical images acquired from modalities such as X-ray, CT, MRI, and ultrasound  \cite{Xiang2025-hy}, thereby supporting many clinical workflows like diagnosis, treatment planning and guidance. Core tasks in this domain include segmentation \cite{Gao2025-jf}, which delineates anatomical structures; registration \cite{Darzi2024-zp}, which aligns images across time, modalities, or subjects; shape completion \cite{friedrich2023point,gafencu2025shape}, which estimates missing or occluded anatomical structures from partial medical observations, and image-to-image translation \cite{Chen2025-cw}, which maps images between different domains, aiming for synthesis or quality enhancement.

With recent advances in deep learning, these tasks are now largely solved, or at least extensively exploited, using data-driven approaches \cite{Liu2021-jv,El_hadramy2023-la}. Deep neural networks have become the dominant tools in medical image computing, achieving state-of-the-art performance across a wide range of benchmarks \cite{Suganyadevi2022-ht}. Despite their success, these methods share a common and often implicit assumption: medical images are represented and processed as discrete grid-based signals. Indeed, in the prevailing deep learning framework, medical images are represented as pixels or voxels arranged on regular lattices and fed directly into neural networks \cite{Suganyadevi2022-ht}. While this representation has proven highly effective, it raises a fundamental question: \textit{is the discrete grid the most appropriate representation for medical images?} This question is particularly relevant given that medical images originate from inherently continuous physical domains but are frequently resampled, interpolated, and transformed during clinical workflows, making discrete grid-based neural network models sensitive to voxel spacing, interpolation, and resampling artifacts.
A continuous formulation would enable resolution-independent inference, improving stability under geometric transformations and outcome consistency across heterogeneous acquisition protocols.

In this paper, we challenge the prevailing deep learning paradigm in medical image computing that learns over discrete image grids and introduce \textbf{NOIR}: \textbf{N}eural \textbf{O}perator mapping for \textbf{I}mplicit \textbf{R}epresentations, a general framework that reframes medical image computing downstream tasks as a continuous operator learning problem. 
%NOIR provides continuous representations of signals, through Implicit Neural Representations (INRs), and uses Neural Operator (NO) to solve downstream tasks. 
Our contributions are as follows:
\begin{itemize}
\item We challenge the standard grid-based deep learning framework in medical image analysis by questioning the choice of image representation.
\item We formulate downstream medical image computing tasks as  Neural Operator (NO) learning problems mapping Implicit Neural Representations (INRs), rather than as discrete grid-based prediction tasks.
\item We demonstrate the applicability of NOIR to core medical imaging tasks, including segmentation, shape completion, image synthesis and image-to-image translation, on several publicly available datasets in both 2D and 3D.
\item We empirically show that NOIR fulfills the fundamental theoretical criteria of neural operators introduced by Kovachki \textit{et al.} \cite{Kovachki2021-ch}, including resolution invariance and mappings between function spaces. Furthermore, we empirically show that NOIR approximates a Representation equivalent NO ($\varepsilon$-ReNO) proposed by Bartolucci \textit{et al.} \cite{Bartolucci2023-hv}.
%\item We show that the learned implicit representations are task-agnostic, enabling the same continuous representation to be reused across multiple downstream tasks without retraining the INRs.

\end{itemize}

%\begin{itemize}
%    \item We challenge the standard grid-based deep learning framework in medical image analysis by questioning the choice of image representation.
%    \item We introduce a novel neural operator framework that operates on INRs of the medical signals.
%    \item We demonstrate that the learned implicit representations are task-agnostic, enabling the same continuous representation to be reused across multiple downstream tasks without retraining the INRs.
%    \item We demonstrate how NOIR can be applied to core medical imaging tasks such as segmentation, registration, and image-to-image translation in several publicly available datasets in both 2D and 3D.
%    \item We empirically validate that NOIR satisfies the theoretical properties of neural operators as defined by Kovachki \textit{et al.} \cite{Kovachki2021-ch}, including resolution invariance and the ability to learn mappings between function spaces.
%\end{itemize}

\section{Related work}

\subsection{Alternative Representations to Medical Image Computing}
In medical image computing, state-of-the-art approaches predominantly rely on convolutional and attention-based architectures. Segmentation methods are mainly based on U-Net variants \cite{Ronneberger2015-ql,Fabian2018-ay,Chen2021-ew,Hu2021-et,Oktay2018-hr} and Vision Transformers adapted to medical imaging \cite{Dosovitskiy2020-ig}. Image-to-image translation is dominated by generative models \cite{Cai2023-xb,Baldini2023-oy}, with recent extensions to diffusion models \cite{Ho2020-dq,Friedrich2024-pu,Durrer2025-id}, while deformable image registration is commonly addressed by dense displacement (or velocity) fields methods \cite{Balakrishnan2018-ef, Chen2021-ew, tian2023gradicon}.
These methods process medical signals as discrete grids represented by pixels or voxels arranged in a regular structure.
%In contrast, the underlying anatomical structures are inherently continuous entities with complex geometries. This motivates the use of continuous representations for medical imaging, which aim to decouple learning from fixed discretizations.
Several alternative methods have explored representations that avoid learning from fixed discretizations.
In MRI, multiple studies have shown that downstream tasks can be performed directly in k-space, enabling lesion detection, classification, and even segmentation without explicit image reconstruction \cite{Rempe2024-ks}. These works demonstrate that diagnostically relevant information is preserved in the Fourier-domain signal and that reconstruction is not a prerequisite for semantic inference \cite{schlemper2017deep,li2024classification,li2024novel,rempe2024k,friedrich2024cwdm}. More broadly, signal-domain learning has also been investigated in other modalities by operating directly on raw acquisition data, such as sinograms in CT and radio-frequency signals in ultrasound, supporting the view that learning can be aligned with the continuous physics of data acquisition rather than discrete image lattices \cite{Zhu2018_AUTOMAP,han2018framing,khan2020adaptive}.
In parallel, functional map representations have been explored for 3D-3D registration by mapping functions between shapes derived from medical images, providing smooth, resolution-independent alternatives to voxel-based registration \cite{litany2017deep,li2025fm}. Together, these approaches highlight viable alternatives to grid-based representations that operate on continuous domains, thereby providing a natural context for INRs.

\subsection{Modeling Images with Implicit Neural Representations}
INRs \cite{Sitzmann2020-yx} model images, shapes, and physical fields as continuous functions rather than discrete arrays, using coordinate-based neural networks that map spatial or spatiotemporal locations to signal values, enabling resolution-independent and functional representations of data \cite{Genova2019-nx, Molaei2023-aa, Sitzmann2020-yx}.
Numerous works focused on improving their expressiveness and training stability, including sinusoidal activation functions \cite{Sitzmann2020-yx}, Fourier feature mappings \cite{Tancik2020-xl}, and multiresolution hash encodings \cite{Muller2022-mf}. This enabled state-of-the-art performance in neural rendering \cite{Mildenhall2020-ng}, 3D reconstruction \cite{Fan2024-zm}, and signal compression \cite{Dupont2022-qm}. However, standard INRs are trained per instance, which limits their generalizability and necessitates costly retraining for each new signal. To address this limitation, recent work has focused on improving the generalizability of INRs through dataset-level learning. Hypernetwork-based \cite{Ha2016-vc} methods use an auxiliary network that conditions INR over the signal representation \cite{Sitzmann2020-dy}. Closely related auto-decoder formulations learn a shared INR together with instance-specific latent codes, forming a continuous latent manifold that supports interpolation and generative modeling \cite{Park2019-kp}, though often at the cost of test-time latent optimization. A key research direction focuses on reducing the reliance on per-instance optimization by learning representations capable of rapid adaptation. Using meta-learning techniques, these approaches train a shared network so that new instances can be fitted at test time with only a few gradient steps, lowering computational costs while enhancing both efficiency and stability when generalizing to unseen signals \cite{Dupont2022-qm, friedrich2025medfuncta, Kazerouni2025-lb}.
Despite these advances, existing INR-based methods primarily focus on instance-level signal representation\cite{Vyas2025-ez,Hadramy2026-gb}, rather than modeling task-specific mappings or transformations between functions, a gap that has motivated the development of NOs for learning mappings between continuous functions. 
%To date, prior work in medical image computing has primarily used INRs to represent images themselves, rather than the image transformations induced by downstream processing tasks such as segmentation, registration, completion, or cross-modal translation.

\subsection{Neural Operators}
NOs have recently emerged as a powerful class of models for learning between infinite-dimensional function spaces by operating directly on functions as inputs and outputs, focusing on the underlying functional relationship rather than on specific discretizations, as described by Kovachki \textit{et al.} \cite{Kovachki2021-ch}.
NOs were initially proposed and have been extensively applied to solve PDEs and physics-based problems. Various NO architectures have since been developed, such as DeepONet \cite{Lu2019-tv}, Fourier Neural Operator (FNO) \cite{Li2020-ss}, Convolutional Neural Operator (CNO) \cite{Raonic_undated-gw}, and Graph Neural Operators (GNO) \cite{Li2020-nb}. The performance of these methods can be sensitive to training conditions such as the grid resolution, domain geometry, and number of discretization points. In particular, the FNO is restricted to regular grids because it relies on spectral convolutions in Fourier space, thereby limiting its applicability to problems with complex or irregular geometries. To overcome these constraints, CORAL \cite{Serrano2023-jl} leverages INRs within an operator learning framework, enabling resolution-invariant function mappings. Overall, these NO methods have been extensively validated on canonical PDEs, including Burgers' equation, Darcy flow, Navier-Stokes equations, and elasticity problems, where they consistently surpass classical neural networks in accuracy. 
However, despite their success in scientific computing, NOs have been sparsely explored in (medical) image computing and are limited to FNO-based methods like FNO3DSeg \cite{Wong2023-gk} and FNOReg \cite{Drozdov2025-br}, which are restricted to regular grids and introduce aliasing errors when processing high-frequency features \cite{Serrano2023-jl}.
Furthermore, they model images as continuous signals without addressing task-level transformations or learn operators without explicit implicit representations.
%Existing approaches either model medical images as continuous signals without addressing task-level transformations, or learn operators without explicit implicit representations.

\section{Methods}

\subsection{Problem Formulation}
\label{sec:problem}

Let $\mathcal{X} \subset \mathbb{R}^d$ denote a continuous spatial domain. We consider supervised learning of mappings between continuous functions defined on $\mathcal{X}$. Given a dataset 
$\mathcal{D} = \{(f_{i}^{d}, g_{i}^{d})\}_{i=1}^N$, each input--output pair consists of discontinuous representations 
\[ 
f_{i}^{d} : \mathcal{X} \to \mathbb{R}^{c_{\mathrm{in}}}, \qquad g_{i}^{d} : \mathcal{X} \to \mathbb{R}^{c_{\mathrm{out}}}, 
\] 
where $f_{i}^{d}$ represents an input discrete signal and $g_{i}^{d}$ denotes a target discrete signal associated with a downstream task. We first seek continuous representations, denoted $\{f_i, g_i\}$, of any given sample $\{f_{i}^{d}, g_{i}^{d}\}$ of the dataset  $\mathcal{D}$, embedding them into suitable function spaces $\mathcal{F}$ and $\mathcal{G}$ respectively, where $\mathcal{F}$ and $\mathcal{G}$ denote spaces of continuous functions on $\mathcal{X}$. 
Subsequently, we aim to learn an operator: $\mathcal{T} : \mathcal{F} \rightarrow \mathcal{G}$,
that maps between these function spaces such that for any given $i < N$, $\mathcal{L}(\mathcal{T}(f_i), g_i) \approx 0$, where $\mathcal{L}$ is a criterion related to the downstream task. No assumptions are made regarding the semantic nature of $g_i$ or the specific form of $\mathcal{L}$, allowing the formulation to remain task-agnostic.

%\subsection{NOIR}
%\label{sec:noir}

%We introduce \textbf{NOIR} (\textbf{N}eural \textbf{O}perator mapping for \textbf{I}mplicit \textbf{R}epresentations), a framework for function space mapping in medical imaging. 

Our proposed framework, NOIR, addresses this problem by learning mappings between function spaces in medical imaging. An overview of the framework is shown in Fig.~\ref{fig:noir_overview}. NOIR contains two main blocks: \textbf{first}, the continuous function representation for discrete medical signals, which constructs implicit representations of the sampled input (in blue) and output (in red) functions, \textbf{second}, the NO mapping that learns to transform between the resulting function spaces. In the following, we describe both blocks in detail.

\begin{figure}[t]
    \centering
    \includegraphics[width=0.70\linewidth]{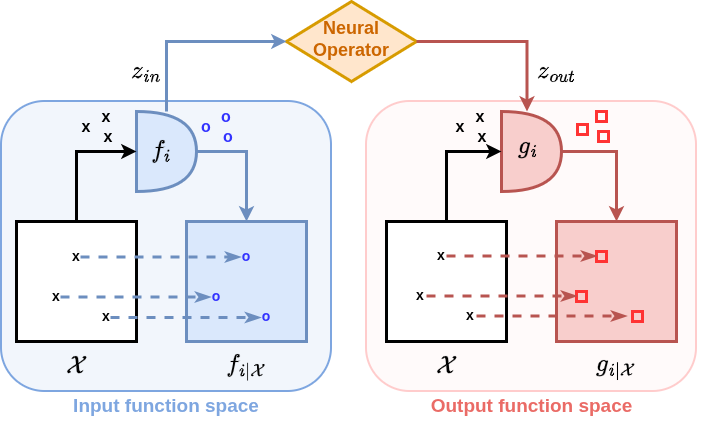}
    \caption{\textbf{Architecture of NOIR}. Discrete input ($f^d_{i}$) and output ($g^d_{i}$) signals are embedded as continuous functions ($f_{i|\mathcal{X}}$ and $g_{i|\mathcal{X}}$) using INRs with shared dataset-level parameters and signal-specific modulations. A neural operator maps the modulations of the input ($z_{in}$) and output ($z_{out}$) INRs, resulting in a continuous function-to-function mapping applicable to multiple downstream tasks.}
    \label{fig:noir_overview}
\end{figure}

\subsection{Continuous Function Representation}
\label{sec:inr}

Given a discrete signal $f_i^d \in \mathcal{D}$, we seek a continuous function $f_i \in \mathcal{F}$ such that $f_{i|\mathcal{X}} \approx f_i^d$. Following \cite{Dupont2022-kn, friedrich2025medfuncta, Serrano2023-jl}, we model $f_i$ with an INR
$\phi_i(x;\theta,\gamma_i)$, where $\theta$ are dataset-level shared parameters and $\gamma_i$ are signal-specific parameters. Instead of optimizing $\gamma_i$ directly, we predict them from optimized latent representations via an auto-decoder. For each sample, a latent vector $z_i \in \mathbb{R}^{p_{\mathrm{in}}}$ is mapped to modulation parameters through a hypernetwork $M_\psi$, i.e., $\gamma_i = M_\psi(z_i)$. These parameters modulate the shared weights $\theta$, adapting a dataset-level learned representation to each sample. This decomposition reflects the strong structural regularities in medical imaging: common content (e.g., background, anatomy, acquisition characteristics) is captured by shared parameters $\theta$, while $\gamma_i$ encodes signal-specific deviations such as anatomical variation or pathology. The resulting continuous approximation is $f_i(x) = \phi_i\big(x;\theta,M_\psi(z_i)\big) \approx f_i^d(x)$.

Training follows a meta-learning scheme (see Algorithm~\ref{alg:train}). The shared parameters $(\theta, \psi)$, governing the INR and the modulation network $M_\psi$, respectively, are optimized in the outer loop across all training signals. For each signal $f_i^d$, an inner loop optimizes a signal-specific latent representation $z_i$, initialized to zero, through $K$ gradient steps minimizing $\mathcal{L}_i$, a similarity loss between the continuous representation $\phi(x;\theta, M_\psi(z_i))$ and its discrete counterpart $f_i^d$. The resulting representation is then used to update $(\theta, \psi)$ via a single outer gradient step, allowing easy adaptation of the shared representation to new signals through low-dimensional latent optimization. At test time (see Algorithm~\ref{alg:test}), $(\theta, \psi)$ are frozen and only the latent representation $z^*$ of the new signal $f^*$ is optimized through the same $K$-step inner loop, yielding a continuous approximation $\phi(x; \theta, M_\psi(z^*))$ without any retraining of the shared parameters.

\begin{figure}[t]
\centering
\begin{minipage}[t]{0.48\textwidth}
\begin{algorithm}[H]
\caption{Training}
\label{alg:train}
\begin{algorithmic}[1]
\STATE \textbf{Input:} dataset $\{f_i^d\}_{i=1}^N$, $\theta$, $M_\psi$
\WHILE{not converged}
    \STATE Sample minibatch $\mathcal{B} \subset \{1,\dots,N\}$
    \STATE $z_i \leftarrow 0,\ \forall i \in \mathcal{B}$
    \FOR{each $i \in \mathcal{B}$}
        \FOR{$k = 1,\dots,K$}
            \STATE $z_i \leftarrow z_i - \alpha\, \nabla_{z_i}\mathcal{L}_i$
        \ENDFOR
    \ENDFOR
    \STATE $(\theta, \psi) \leftarrow (\theta, \psi) - \beta\, \nabla_{(\theta,\psi)}\textstyle\sum_{i \in \mathcal{B}} \mathcal{L}_i$
\ENDWHILE
\STATE \textbf{Return:} trained $\theta$, $\psi$
\end{algorithmic}
\end{algorithm}
\end{minipage}
\hfill
\begin{minipage}[t]{0.48\textwidth}
\begin{algorithm}[H]
\caption{Fitting new signal}
\label{alg:test}
\begin{algorithmic}[1]
\STATE \textbf{Input:} new signal $f^*$, trained $(\theta, \psi)$
\STATE $z^* \leftarrow 0$
\FOR{$k = 1,\dots,K$}
    \STATE $z^* \leftarrow z^* - \alpha\, \nabla_{z^*}\mathcal{L}(f^*, \phi_{\theta,\psi}(z^*))$
\ENDFOR
\STATE \textbf{Return:} $\phi\big(\textbf{.};\theta, M_\psi(z^*)\big)$
\end{algorithmic}
\end{algorithm}
\end{minipage}
\end{figure}

\subsection{Neural Operator Mapping}
\label{sec:no}
Following the problem formulation of Section~\ref{sec:problem}, we operate in a supervised learning setting with a dataset $\mathcal{D} = \{(f_i^d, g_i^d)\}_{i=1}^N$ of paired discrete signal representations. We first apply the continuous function approximation described above to both signals of each pair, obtaining their respective continuous representations as modulated INRs. Specifically, let $z_{\text{in}}^i$ and $z_{\text{out}}^i$ denote the latent representations encoding the continuous functions $f_i^d$ and $g_i^d$, respectively, obtained by running Algorithm~\ref{alg:test}. We learn a NO that maps input modulations to output modulations, i.e., $\hat{z}_{\text{out}} = \mathcal{T}(z_{\text{in}})$, where $\mathcal{T}$ denotes any data-driven parametric model capable of learning the mapping from $z_{\text{in}}^i$ to $z_{\text{out}}^i$. In our experiments, we instantiate $\mathcal{T}$ as a multi-layer perceptron (MLP) with residual connections, which we find sufficient to model this transformation. The learning of $\mathcal{T}$ is supervised by minimizing the mean squared error between the predicted and ground truth latent representations $\mathcal{L}_{\text{NO}} = \frac{1}{N}\sum_{i=1}^{N} \left\| \mathcal{T}(z_{\text{in}}^i) - z_{\text{out}}^i \right\|_{2}^2 $.

%supervising the neural operator. 
% We empirically show in the results section that this formulation respects the definition of a NO as defined by Kovachki \textit{et al.}  \cite{Kovachki2021-ch} in terms of resolution invariance.

\subsection{NOIR is an $\epsilon$-ReNO}
\label{sec:reno}

NOs must maintain consistency between continuous function spaces and their discrete representations to avoid aliasing errors that compromise performance across different resolutions~\cite{Kovachki2021-ch}. Following the framework of Bartolucci \textit{et al.} \cite{Bartolucci2023-hv}, we define a \textit{Representation equivalent Neural Operator} (ReNO) as a neural operator $(\mathcal{T}, \mathcal{T}^d)$ for which the aliasing error $\varepsilon(\mathcal{T}, \mathcal{T}^d, \Psi, \Phi) = 0$ for all admissible frame sequence pairs $(\Psi, \Phi)$. Here, $\Psi = \{\psi_i\}_{i \in I}$ and $\Phi = \{\phi_k\}_{k \in K}$ are frame sequences for the input and output function spaces $\mathcal{F}$ and $\mathcal{G}$, respectively, with synthesis operators $T_\Psi$ and $T_\Phi$ that map discrete coefficients to continuous functions, and analysis operators $T^\dagger_\Psi$ and $T^\dagger_\Phi$ that extract discrete representations from continuous functions. The aliasing error operator measures the discrepancy between the continuous operator $\mathcal{T}$ and its discrete representation $\mathcal{T}^d$ after synthesis and analysis:
\begin{equation}
\varepsilon(\mathcal{T}, \mathcal{T}^d, \Psi, \Phi) = \mathcal{T} - T_\Phi \circ \mathcal{T}^d \circ T^\dagger_\Psi.
\end{equation}
When this error is bounded by a small constant $\epsilon$, i.e., $\|\varepsilon(\mathcal{T}, \mathcal{T}^d)\| \leq \epsilon$, the operator is termed an $\epsilon$-ReNO, and the representation equivalence error between different discretizations satisfies $\|\tau(\mathcal{T}^d, \mathcal{T}'^d)\| \leq 2\epsilon\sqrt{B_\Psi}\sqrt{A_\Phi}$, where $A_\Phi, B_\Psi$ are frame bounds and $\tau(\mathcal{T}^d, \mathcal{T}'^d) = \mathcal{T}^d - T^\dagger_\Phi \circ T_{\Phi'} \circ \mathcal{T}'^d \circ T^\dagger_{\Psi'} \circ T_\Psi$ quantifies the consistency between discrete operators $\mathcal{T}^d$ and $\mathcal{T}'^d$ associated with different frame sequences. In \textbf{NOIR}, the frame sequences $\Psi$ and $\Phi$ correspond to different spatial discretizations of the input signals $f_i^d$ and output signals $g_i^d$ at resolutions $\mathcal{X}^{(r)} \subset \mathcal{X}$ with $r$ a set of resolutions. The synthesis operator $T_\Psi$ maps discrete point samples to continuous functions via the input INR $\phi_{\text{in}}(\cdot; \theta_{\text{in}}, M_{\psi_{\text{in}}}(z_{\text{in}}))$, while the analysis operator $T^\dagger_\Psi$ extracts the discrete latent representation $z_{\text{in}} \in \mathbb{R}^{p_{\mathrm{in}}}$ through the inner optimization loop of Algorithm~\ref{alg:test}. Similarly, $T_\Phi$ and $T^\dagger_\Phi$ operate through the output INR $\phi_{\text{out}}(\cdot; \theta_{\text{out}}, M_{\psi_{\text{out}}}(z_{\text{out}}))$ with latent representations $z_{\text{out}} \in \mathbb{R}^{p_{\mathrm{out}}}$. The continuous operator $\mathcal{T}: \mathcal{F} \to \mathcal{G}$ maps between function spaces, while its discrete counterpart $\mathcal{T}^d: \mathbb{R}^{p_{\mathrm{in}}} \to \mathbb{R}^{p_{\mathrm{out}}}$ operates on latent representations. The complete pipeline becomes $T_\Phi \circ \mathcal{T}^d \circ T^\dagger_\Psi$, where $T^\dagger_\Psi$ optimizes the input latent representation $z_{\text{in}}$ from discrete observations at resolution $r$, $\mathcal{T}^d$ predicts the output latent representation $\hat{z}_{\text{out}} = \mathcal{T}^d(z_{\text{in}})$, and $T_\Phi$ reconstructs the continuous output function via the output INR. 
We provide in Section~\ref{eps-ReNO-experiments} empirical evidence to validate that NOIR constitutes an $\epsilon$-ReNO.
\section{Experiments}

\subsection{Datasets and implementation details}
To demonstrate the effectiveness of NOIR, we evaluated our method on the following medical image computing tasks: segmentation, shape completion, image translation, and synthesis. Experiments were conducted using four public datasets: Shenzhen \cite{Jaeger2014-kh}, OASIS-4 \cite{Marcus2007-or}, SkullBreak \cite{Kodym2021-xv}, fastMRI \cite{Knoll2020-iu}, along with one in-house clinical dataset for ultrasound synthesis. Each dataset was divided into training, validation, and test splits, and all reported results were obtained using the held-out test sets after model selection based on validation performance. 
For segmentation and anatomy completion, we evaluated the performance using the Dice Similarity Coefficient (DSC) and Intersection over Union (IoU), while image translation and synthesis were assessed using Peak Signal-to-Noise Ratio (PSNR), Structural Similarity Index (SSIM). 
For each task, we first trained separate INRs for the input and output domains on their respective datasets, and subsequently trained the NO to learn the mapping between them.
All INR layers used sine activation functions except for the final layer, which used a sigmoid for the input INR and a softmax to produce class probabilities for the output INR.
Across tasks, the output INR differs only in the dimensionality of its final layer, which is adapted to the specific prediction setting. All experiments were conducted on a single NVIDIA A100 GPU (40\,GB). The INRs were trained with inner- and outer-loop learning rates of $1 \times 10^{-2}$ and $5 \times 10^{-6}$, respectively, using a batch size of 1 and 10{,}000 sampled points per iteration. The inner loop was run for 5 iterations during training and 10 iterations at test time. The NOs were trained separately with a learning rate of $5 \times 10^{-6}$. All models employed early stopping with a patience of 50 epochs and a maximum of 1000 training epochs.

%We compare NOIR against several methods that represent different architectural paradigms in medical image computing: U-Net \cite{Ronneberger2015-ql} (CNN-based), Vision Transformer (ViT) \cite{Dosovitskiy2020-ig} (attention-based), Attention U-Net (AttU-Net) \cite{Oktay2018-hr} (CNN and attention-based), Fourier Neural Operator (FNO)\cite{Li2020-ss} (spectral operator), and Denoising Diffusion Probabilistic Models (DDPM) \cite{Ho2020-dq}, a diffusion model for synthesis and translation tasks. 

\subsection{Baseline methods}
We compared NOIR against several methods that represent different architectural paradigms in medical image computing: U-Net \cite{Ronneberger2015-ql}, a convolutional encoder–decoder baseline; ViT \cite{Dosovitskiy2020-ig}, an attention-based model operating on tokenized image patches; Attention U-Net (AttU-Net) \cite{Oktay2018-hr}, which augments convolutional features with attention gating; FNO \cite{Li2020-ss}, a spectral operator-learning framework; and Denoising Diffusion Probabilistic Models (DDPM) \cite{Ho2020-dq}.
These methods have demonstrated strong performance in grid-based medical image analysis, and we aim to investigate whether NOIR's formulation can achieve comparable or better performance at the native training and unseen resolutions. We chose the \textit{vanilla} architecture of these models and trained them on identical datasets and splits, with more details in the supplementary material. In the following sections, we present the results for each of the downstream tasks.

\setlength{\aboverulesep}{1pt}
\setlength{\belowrulesep}{1pt}
\renewcommand{\arraystretch}{0.7}
{
\begin{table}[tb]
  \caption{Performance of NOIR and baseline methods across segmentation and anatomy completion tasks. }
  \label{tab:results1}
  \centering
  \begin{tabular}{@{}llcccccc@{}}
    \toprule
    \multirow{2}{*}{Dataset} &
    \multirow{2}{*}{Method} &
    \multicolumn{2}{c}{Res = $10\%$} &
    \multicolumn{2}{c}{Res = $25\%$} &
    \multicolumn{2}{c}{Res = $100\%$} \\
    \cmidrule(lr){3-4} \cmidrule(lr){5-6} \cmidrule(lr){7-8}
    & & DSC $\uparrow$ & IoU $\uparrow$ & DSC $\uparrow$ & IoU $\uparrow$ & DSC $\uparrow$ & IoU $\uparrow$ \\
    \midrule

    \multirow{4}{*}{Shenzhen \cite{Jaeger2014-kh}}
    & U-Net \cite{Ronneberger2015-ql}            & 0.75 & 0.62 & 0.92 & 0.87 & \textbf{0.95} & \textbf{0.91} \\
    & ViT  \cite{Dosovitskiy2020-ig}             & 0.91 & 0.84 & \underline{0.93} & \underline{0.87} & \underline{0.94} & \underline{0.90} \\
    & AttU-Net \cite{Oktay2018-hr}               & 0.01 & 0.01 & 0.74 & 0.64 & \textbf{0.95 }& \textbf{0.91} \\
    & FNO \cite{Li2020-ss}                       & \underline{0.92} & \underline{0.86} & \textbf{0.94} & \textbf{0.88} & \underline{0.94} & 0.89 \\
    & \textbf{NOIR (Ours)}                       & \textbf{0.93} & \textbf{0.88} & \textbf{0.94} & \textbf{0.88} & \underline{0.94} & 0.88 \\
    \midrule

    \multirow{4}{*}{OASIS-4 \cite{Marcus2007-or}}
    & U-Net \cite{Ronneberger2015-ql}            & 0.40 & 0.29 & 0.79 & 0.66 & \textbf{0.95} & \textbf{0.90} \\
    & ViT \cite{Dosovitskiy2020-ig}              & \underline{0.66} & \underline{0.51} & \textbf{0.85} & \underline{0.74} & 0.90 & \underline{0.83} \\
    & AttU-Net \cite{Oktay2018-hr}               & 0.40 & 0.29 & 0.76 & 0.64 & \textbf{0.95} & \textbf{0.90} \\
    & FNO \cite{Li2020-ss}                       & 0.65 & \underline{0.51} & \underline{0.84} & \underline{0.74} & \underline{0.91} & 0.83 \\
    & \textbf{NOIR (Ours)}                       & \textbf{0.85} & \textbf{0.79} & \textbf{0.85} & \textbf{0.79} & 0.85 & 0.79 \\
    \midrule

    % \multirow{4}{*}{OASIS-24 \cite{Marcus2007-or}}
    % & U-Net \cite{Ronneberger2015-ql}            & 0.20 & 0.14 & 0.68 & 0.57 & \textbf{0.86} & \textbf{0.79} \\
    % & ViT \cite{Dosovitskiy2020-ig}              & \underline{0.54} & \underline{0.44} & 0.70 & 0.59 & 0.75 & 0.65 \\
    % & AttU-Net \cite{Oktay2018-hr}               & 0.25 & 0.18 & 0.68 & 0.58 & \textbf{0.86} & \textbf{0.79} \\
    % & FNO \cite{Li2020-ss}                       & 0.52 & 0.40 & \underline{0.71} & \underline{0.60} & \underline{0.83} & \underline{0.75} \\
    % & \textbf{NOIR (Ours)}                       & \textbf{0.74} & \textbf{0.63} & \textbf{0.74} & \textbf{0.63} & 0.74 & 0.63 \\
    % \midrule

    \multirow{3}{*}{SkullBreak \cite{Kodym2021-xv}}
    & 3D U-Net \cite{Ronneberger2015-ql}         & 0.71 & 0.56 & \underline{0.90} & \underline{0.83} & \textbf{0.96} & \textbf{0.93} \\
    & 3D ViT \cite{Dosovitskiy2020-ig}           & \textbf{0.86} & \textbf{0.76} & \textbf{0.91} & \textbf{0.84} & \underline{0.91} & \underline{0.84} \\
    & 3D AttUNet \cite{Oktay2018-hr}             & 0.52 & 0.35 & 0.84 & 0.73 & \textbf{0.96} & 0\textbf{.93} \\
    & FNO \cite{Li2020-ss} & \underline{0.74} & 0.59 & 0.87 & 0.78 & 0.90 & 0.82 \\
    & \textbf{NOIR (Ours)}                       & \textbf{0.86} & \underline{0.73} & 0.86 & 0.75 & 0.87 & 0.75 \\
    \bottomrule
  \end{tabular}
\end{table}
}

\subsection{NOIR for image segmentation}
\begin{figure}[t]
    \centering
    \includegraphics[width=0.9\linewidth]{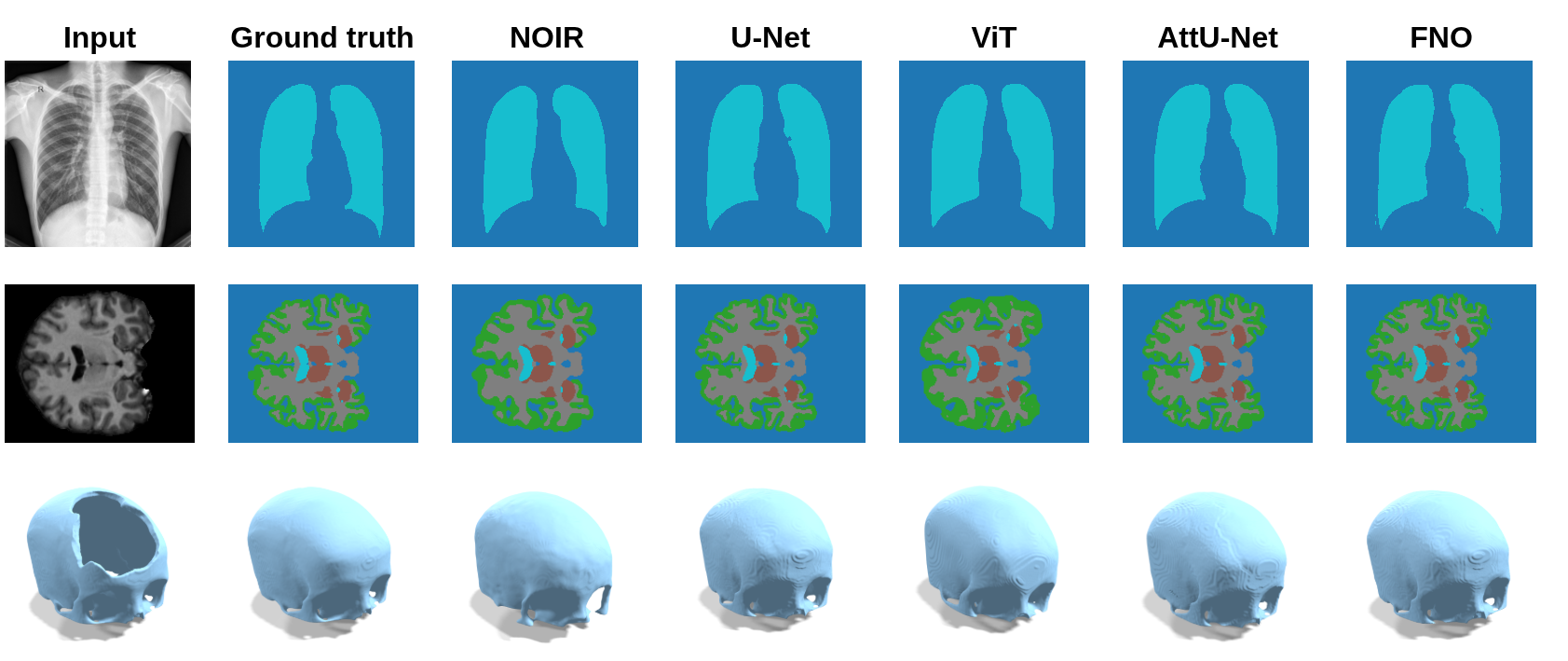}
    \caption{Qualitative results for segmentation and anatomy completion. Rows correspond to Shenzhen binary segmentation, OASIS-4 multi-label segmentation, and SkullBreak anatomy completion, respectively. A randomly selected test sample is shown with the input image, ground truth, NOIR results, and comparison methods.}
    \label{fig:segmentation}
\end{figure}

For the image segmentation task, we present two representative settings: binary segmentation using the Shenzhen chest X-ray dataset and multi-label segmentation using the OASIS-4 brain MRI dataset. The Shenzhen \cite{Jaeger2014-kh} dataset contains 662 samples, which are split into training, validation, and test sets with ratios of 0.8, 0.1, and 0.1, respectively. The images were downsampled to $200 \times 200$. In contrast, OASIS-4 \cite{Marcus2007-or} brain MRI 2D slices of size $160 \times 192$ and multi-label anatomical annotations. The OASIS-4 dataset contains 457 samples, which are split into training, validation, and test sets using ratios of 0.7, 0.15, and 0.15. In both experiments, the input INR shares the same architecture, with modulation vectors of size 64 for Shenzhen and 2048 for OASIS-4. The output INR uses a final layer of size 2 for Shenzhen, corresponding to binary segmentation, and size 5 for OASIS-4, corresponding to four anatomical classes plus background.
The NO maps between modulation spaces using a three-layer fully connected architecture with 512 hidden units per layer. Table~\ref{tab:results1} presents the segmentation performance of NOIR on the test split for both datasets at 10\%, 25\% and 100\% (full) resolution.
In Figure~\ref{fig:segmentation}, we present qualitative results comparing NOIR with the baseline methods over one random test sample.

\subsection{NOIR for shape completion}
For the shape completion task, we evaluated NOIR on the SkullBreak dataset \cite{Kodym2021-xv}, which contained 3D volumes of size $512 \times 512 \times 512$ representing defective skulls along with their corresponding complete skulls. The dataset split used follows the one provided by \cite{Kodym2021-xv}. We trained the input INR on the defective skull volumes and the output INR on the complete skull volumes, using latent codes of size 64 to represent each volume. The NO maps these 64-dimensional modulations with a three-layer fully connected network of 128 hidden units per layer. Table~\ref{tab:results1} presents quantitative results on the test split of NOIR and the relative grid-based comparing methods. In the third row of Figure~\ref{fig:segmentation}, we show qualitative anatomy completion results, comparing NOIR with the baseline methods.

\begin{figure}[t]
    \centering
    \includegraphics[width=\linewidth]{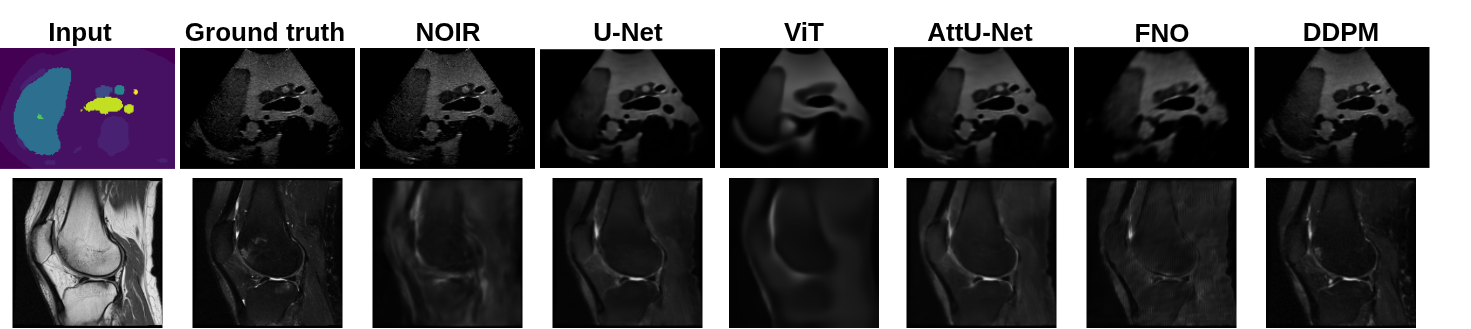}
    \caption{Qualitative results for image synthesis (top row) and image translation (bottom row). A randomly selected test sample is shown, including the input image, ground truth, NOIR results, and results from comparison methods.}
    \label{fig:domain_adaptation}
\end{figure}

\subsection{NOIR for image translation and synthesis}
To demonstrate the ability of NOIR to bridge the gap between different modalities, we considered two tasks: image translation and image synthesis. For image translation, we used the fastMRI \cite{Knoll2020-iu} dataset, which provides paired and registered knee volumes in Proton Density (PD) and T2 modalities. We extracted 2D paired slices from 533 patients, selecting 11 central slices per volume, resulting in 5863 pairs of PD and T2 images. For ultrasound synthesis, we used an in-house private dataset of 333 paired samples acquired on a phantom with a tracked probe, enabling the estimation of ground-truth pose between ultrasound images and CT-derived label maps. In both settings, datasets were split at the patient level into training, validation, and test sets with ratios 0.8, 0.1, and 0.1. Separate INRs were trained for each domain, using modulation vectors of size 4096 for fastMRI and 64 for ultrasound. For fastMRI, the input and output INRs model PD and T2 images, respectively, while for ultrasound synthesis they model label maps and corresponding ultrasound images. The NO was then trained to map between the corresponding modulation spaces, using three hidden layers of 512 units each for the ultrasound task and a single hidden layer of 2048 units for the fastMRI task.
Table~\ref{tab:translation_results} reports quantitative PSNR and SSIM results comparing NOIR with baseline methods, while Figure~\ref{fig:domain_adaptation} presents qualitative comparisons.

\setlength{\aboverulesep}{1pt}
\setlength{\belowrulesep}{1pt}
\renewcommand{\arraystretch}{0.7}
{
\begin{table}[tb]
  \caption{Image-to-image translation performance for domain adaptation and ultrasound synthesis. PSNR and SSIM are reported across different resolutions.}
  \label{tab:translation_results}
  \centering
  \begin{tabular}{@{}llcccccc@{}}
    \toprule
    \multirow{2}{*}{Task} &
    \multirow{2}{*}{Method} &
    \multicolumn{2}{c}{Res = $25\%$} &
    \multicolumn{2}{c}{Res = $50\%$} &
    \multicolumn{2}{c}{Res = $100\%$} \\
    \cmidrule(lr){3-4} \cmidrule(lr){5-6} \cmidrule(lr){7-8}
    & & PSNR $\uparrow$ & SSIM $\uparrow$ & PSNR $\uparrow$ & SSIM $\uparrow$ & PSNR $\uparrow$ & SSIM $\uparrow$ \\
    \midrule

    \multirow{6}{*}{US Synthesis}
    & U-Net \cite{Ronneberger2015-ql}          & 17.51 & 0.84 & 17.45 & 0.84 & 17.54 & \underline{0.89} \\
    & ViT \cite{Dosovitskiy2020-ig}            & \underline{28.22} & 0.78 & \underline{28.31} & 0.79 & 28.36 & 0.79 \\
    & AttU-Net \cite{Oktay2018-hr}             & 27.76 & \textbf{0.87} & 28.29 & \textbf{0.89} & \underline{28.55} & \textbf{0.90} \\
    & FNO \cite{Li2020-ss}                     & 27.37 & 0.73 & 27.94 & 0.75 & 28.15 & 0.76 \\
    & DDPM \cite{Ho2020-dq}                     & 26.35 & 0.65 & 26.51 & 0.65 & 26.91 & 0.64 \\
    & \textbf{NOIR (Ours)}                     & \textbf{30.94} & \underline{0.86} & \textbf{30.94} & \underline{0.86} & \textbf{31.83} & 0.87 \\

    \midrule
    \multirow{6}{*}{\centering FastMRI \cite{Knoll2020-iu}}
    & U-Net \cite{Ronneberger2015-ql}          & 21.68 & 0.60 & \underline{23.49} & \underline{0.68} & \underline{24.55} & \underline{0.72} \\
    & ViT \cite{Dosovitskiy2020-ig}            & 22.34 & 0.62 & 22.54 & 0.63 & 22.61 & 0.63 \\
    & AttU-Net \cite{Oktay2018-hr}             & 21.75 & 0.60 & 23.26 & 0.67 & 24.46 & \underline{0.72} \\
    & FNO \cite{Li2020-ss}                     & 22.61 & 0.58 & 22.74 & 0.60 & 22.82 & 0.60 \\
    & DDPM \cite{Ho2020-dq}                     & 22.66 & \textbf{0.69} & \textbf{25.59} & \textbf{0.76} & \textbf{25.45} & \textbf{0.74} \\
    & \textbf{NOIR (Ours)}                     & \textbf{22.87} & \underline{0.61} & 22.87 & 0.61 & 22.87 & 0.61 \\

    \bottomrule
  \end{tabular}
\end{table}
}

\subsection{Empirical evidence that NOIR is an $\epsilon$-ReNO}
\label{eps-ReNO-experiments}

\begin{figure}[t]
\centering

\begin{subfigure}{0.52\linewidth}
    \centering
    \includegraphics[width=\linewidth]{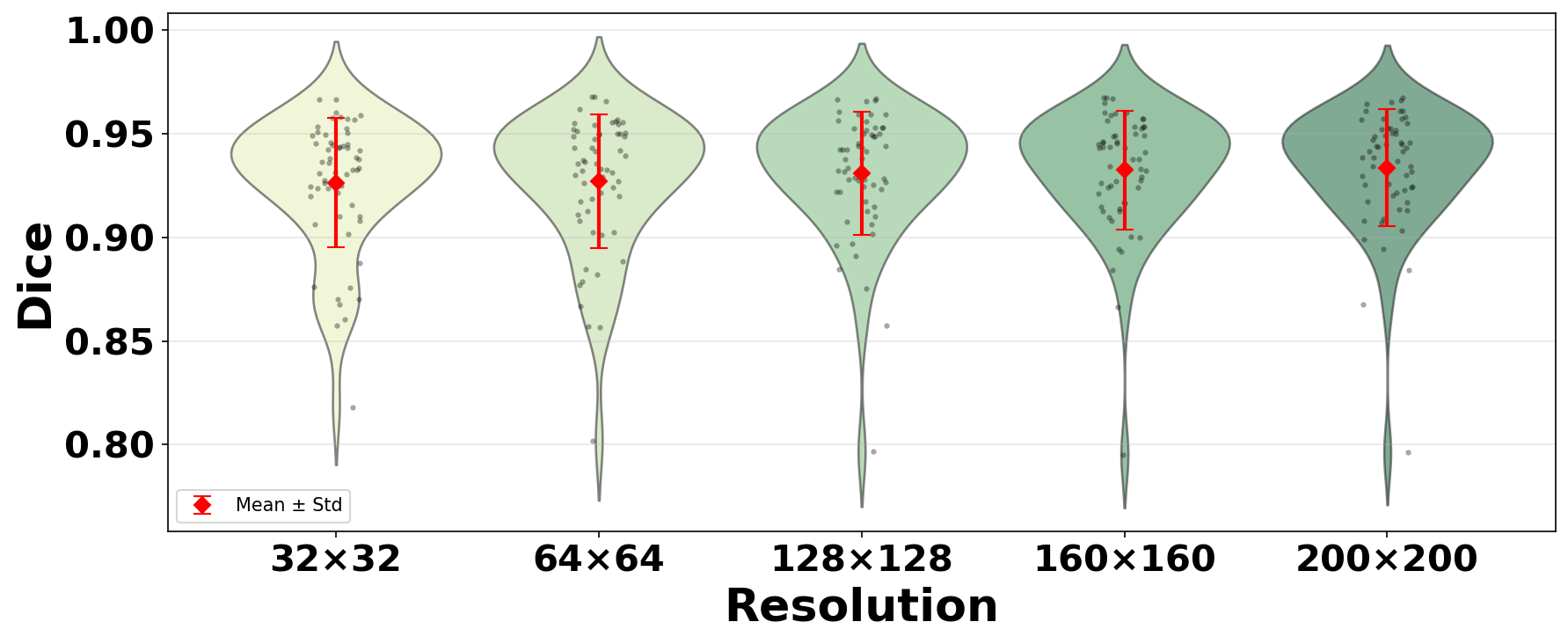}
    \caption{}
    %\caption{Dice score distributions for NOIR evaluated across input resolutions from $32^2$ to $200^2$ on the Shenzhen dataset.}
    \label{fig:dice_distribution}
\end{subfigure}
\hfill
\begin{subfigure}{0.47\linewidth}
    \centering
    \includegraphics[width=\linewidth]{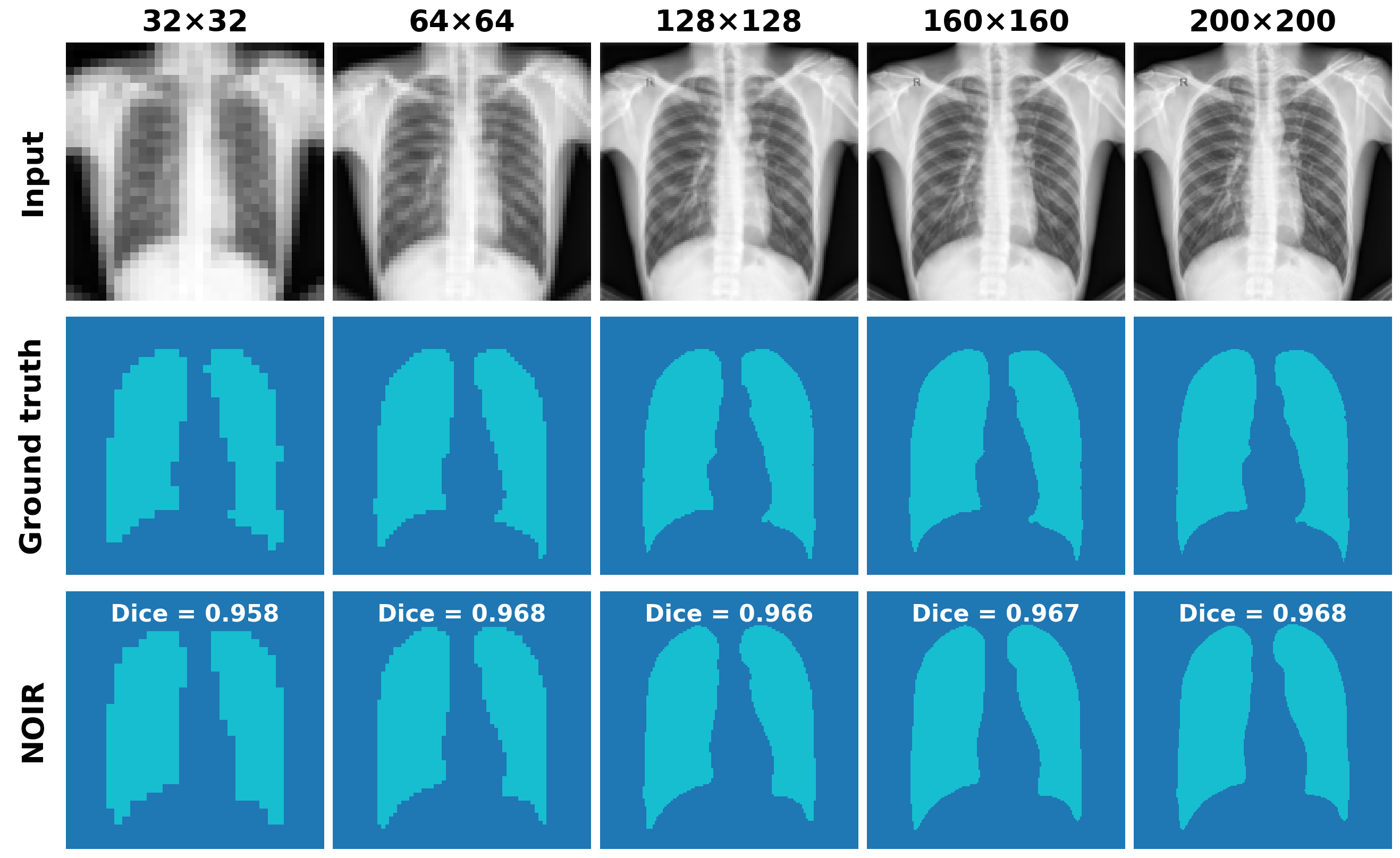}
    \caption{}
    %\caption{Example test sample showing input X-rays at multiple resolutions with ground truth and NOIR predictions.}
    \label{fig:qualitative_epsilon_reno}
\end{subfigure}

\caption{Resolution robustness of NOIR. (a) Dice distributions across discretizations show stable performance with mean Dice around $0.93$--$0.94$. (b) Qualitative example demonstrating consistent segmentation predictions across different input resolutions.}

\end{figure}

To validate that NOIR constitutes an $\epsilon$-ReNO \cite{Bartolucci2023-hv}, we tested the trained model from the Shenzhen experiments across multiple input resolutions: $32^2$, $64^2$, $128^2$, $160^2$, and $200^2$. For each test sample, we optimized the input latent code $z_{\text{in}}^{(r)}$ at each resolution $r$ through the inner loop of Algorithm~\ref{alg:test}, then predicted the corresponding output latent $\hat{z}_{\text{out}}^{(r)} = \mathcal{T}^d(z_{\text{in}}^{(r)})$ via the neural operator. We computed pairwise mean squared errors $\|z_{\text{in}}^{(r_i)} - z_{\text{in}}^{(r_j)}\|_2^2$ and $\|\hat{z}_{\text{out}}^{(r_i)} - \hat{z}_{\text{out}}^{(r_j)}\|_2^2$ across all resolution pairs $(r_i, r_j)$ for each sample. The maximum discrepancy across all pairs for each sample provides a per-sample aliasing bound, and we defined $\epsilon$ as the maximum of these bounds across the entire test set. This yields $\epsilon_{z_{\text{in}}} = 4.0 \times 10^{-6}$ for input latents and $\epsilon_{z_{\text{out}}} = 2.2 \times 10^{-5}$ for predicted output latents, indicating that the neural operator produces highly consistent latent predictions regardless of input discretization. To assess the representation equivalence error $\|\tau(\mathcal{T}^d, \mathcal{T}'^d)\|$ at the segmentation level, we reconstructed segmentations from each predicted latent code at its native resolution $r$ and computed the mean squared error with respect to the corresponding ground truth segmentation at the same resolution. The maximum MSE across all samples and resolutions yielded $\epsilon_{\text{seg}} = 1.8 \times 10^{-2}$, confirming that NOIR maintains accurate predictions across different input discretizations with minimal aliasing. Figure~\ref{fig:dice_distribution} presents the distribution of DSCs per resolution. Figure~\ref{fig:qualitative_epsilon_reno} provides a visual example for a representative test sample, showing the input X-ray images at different resolutions alongside their corresponding ground truth segmentation masks and NOIR predictions. 
 %As observed, performance remains remarkably stable across all resolutions, with mean Dice scores ranging from $\approx 0.93$ to $\approx 0.94$ across the entire resolution spectrum from $32^2$ to $200^2$, highlighting the robustness of NOIR to varying input discretizations. 
%, further illustrating the consistency of the learned operator across various discretizations.

\begin{table}[tb]
\centering
\caption{Ablation study on NO model selection.}
\label{tab:ablation_operator}
\rowcolors{2}{}{}
\begin{tabular}{@{}llcc@{}}
\toprule
Dataset & Operator $\mathcal{T}$ & DSC $\uparrow$ & IoU $\uparrow$ \\
\midrule
\multirow{4}{*}{\parbox{1.8cm}{Shenzhen\\($p=64$)}} 
    & Linear Regression & 0.93 & 0.87 \\
    & SVR              & \textbf{0.94} & \textbf{0.88} \\
    & Random Forest    & 0.93 & 0.86 \\
    & \textbf{MLP}     & \textbf{0.94} & \textbf{0.88} \\
\midrule
\multirow{4}{*}{\parbox{1.8cm}{OASIS-4\\($p=2048$)}} 
    & Linear Regression & 0.80 & 0.68 \\
    & SVR              & 0.79 & 0.66 \\
    & Random Forest    & 0.80 & 0.68 \\
    & \textbf{MLP}     & \textbf{0.85} & \textbf{0.79} \\
\bottomrule
\end{tabular}
\end{table}

% \renewcommand{\arraystretch}{0.7}
% \begin{table*}[tb]
% \centering
% \begin{minipage}[t]{0.35\textwidth}
% \centering
% \caption{Segmentation performance comparison.}
% \label{tab:task-agnostic}
% \rowcolors{2}{}{}
% \begin{tabular}{@{}lcc@{}}
% \toprule
% Method & DSC $\uparrow$ & IoU $\uparrow$ \\
% \midrule
% U-Net            & 0.91 & 0.86 \\
% ViT              & 0.89 & 0.83 \\
% AttU-Net         & 0.91 & 0.86 \\
% FNO              & 0.87 & 0.80 \\
% \textbf{NOIR (Ours)} & \textbf{--} & \textbf{--} \\
% \bottomrule
% \end{tabular}
% \end{minipage}
% \hfill
% \begin{minipage}[t]{0.61\textwidth}
% \centering
% \caption{Ablation study on NO model selection.}
% \label{tab:ablation_operator}
% \rowcolors{2}{}{}
% \begin{tabular}{@{}llcc@{}}
% \toprule
% Dataset & Operator $\mathcal{T}$ & DSC $\uparrow$ & IoU $\uparrow$ \\
% \midrule
% \multirow{5}{*}{\parbox{1.8cm}{Shenzhen\\($p=64$)}} 
%     & Linear Regression & 0.93 & 0.87 \\
%     & SVR              & \textbf{0.94} & \textbf{0.88} \\
%     & Random Forest    & 0.93 & 0.86 \\
%     %& XGBoost          & \textbf{0.94} & \textbf{0.88} \\
%     & \textbf{MLP} & \textbf{0.94} & \textbf{0.88} \\
% \midrule
% \multirow{5}{*}{\parbox{1.8cm}{OASIS-4\\($p=2048$)}} 
%     & Linear Regression & 0.80 & 0.68 \\
%     & SVR              & 0.79 & 0.66 \\
%     & Random Forest    & 0.80 & 0.68 \\
%     %& XGBoost          & -- & -- \\
%     & \textbf{MLP } & \textbf{0.85} & \textbf{0.79} \\
% \bottomrule
% \end{tabular}
% \end{minipage}
% \end{table*}
\subsection{Ablation Study}
\label{sec:ablation_study}

While we instantiate the neural operator $\mathcal{T}$ as a MLP with residual connections for mapping latent representations $z_{\text{in}} \to z_{\text{out}}$, this choice is not unique. The optimal model for $\mathcal{T}$ depends on the complexity of the function space mapping, which is influenced by both the dimensionality of the latent spaces ($p_{\mathrm{in}}$, $p_{\mathrm{out}}$) and the size of the training dataset $|\mathcal{D}|$. However, regardless of the specific model chosen for $\mathcal{T}$, NOIR remains a neural operator, since the synthesis operators $T_\Psi$ and $T_\Phi$ and analysis operators $T^\dagger_\Psi$ and $T^\dagger_\Phi$ remain unchanged, ensuring the mapping between continuous function spaces $\mathcal{F} \to \mathcal{G}$. To demonstrate the flexibility and impact of the model selection, we conducted ablation experiments on the segmentation tasks for Shenzhen and OASIS-4, replacing the MLP operator with classical data-driven models: Linear Regression, Support Vector Regression (SVR), and Random Forest. For Shenzhen, the latent dimensions are relatively small ($p_{\mathrm{in}} = p_{\mathrm{out}} = 64$), whereas OASIS-4 employs high-dimensional latent spaces ($p_{\mathrm{in}} = p_{\mathrm{out}} = 2048$). As shown in Table~\ref{tab:ablation_operator}, for Shenzhen, all models achieve comparable performance. In contrast, for OASIS-4, the MLP outperforms classical methods, suggesting that higher-dimensional latent mappings benefit from the capacity of deep networks. This supports choosing $\mathcal{T}$ as an MLP, as it flexibly captures both simple and high-dimensional transformations.

\section{Discussion}

The experiments demonstrate that NOIR provides a competitive and conceptually distinct alternative to conventional grid-based deep learning models for medical image computing. Across a diverse set of tasks: segmentation, shape completion, image translation, and synthesis. NOIR consistently achieves performance comparable to or better than established architectures, including U-Net, ViT, AttU-Net, FNO, and DDPM.

In binary \textbf{segmentation}, NOIR achieves a DSC of $0.94$ at full resolution, matching FNO and ViT, while U-Net and AttU-Net remain the top-performing baselines at $0.95$. More importantly, this performance is preserved by NOIR across all sub-training resolutions, where grid-based methods such as U-Net, ViT, and AttU-Net suffer a marked degradation in accuracy. FNO remains relatively stable at lower resolutions, though it exhibits a small but consistent aliasing error on this dataset. On the more challenging Oasis-4 dataset, NOIR outperforms all compared methods at low resolutions and sits within $6\%$ Dice of the best-performing baseline at full resolution. FNO, by contrast, shows an $18\%$ discrepancy between its results at high and lowest resolution, corroborating the susceptibility of spectral methods to aliasing errors as reported by Bartolucci \textit{et al.}~\cite{Bartolucci2023-hv} and others. For \textbf{shape completion}, NOIR successfully completes the missing anatomical structures in most cases, as evidenced by the qualitative results. However, it struggles to recover fine-grained surface details, most notably around the orbital region of the skull, where subtle geometric features present in the input are not fully reproduced in the prediction, see Figure \ref{fig:segmentation}. We believe the performance could be better assessed with a region-of-interest DSC focused on the missing area. In addition, the incorporation of an anatomical consistency loss function during training, or dedicated post-processing, might improve the results. Overall, NOIR achieves competitive performance on this task, with only $3\%$ and $4\%$ Dice below FNO and ViT, respectively, at full resolution. Crucially, its predictions remain stable under resolution degradation, allowing it to outperform all compared methods at the lowest resolution.

For \textbf{image synthesis}, NOIR outperforms all compared methods on the ultrasound generation task in terms of PSNR, while maintaining consistent accuracy across all resolutions. We attribute this to the point-wise nature of INR-based econstruction, which avoids the implicit smoothing introduced by convolutional kernels or the patch-level averaging of attention mechanisms. This effect is visible in the qualitative results (Figure \ref{fig:domain_adaptation}): while U-Net, ViT, and AttU-Net preserve structural consistency between label maps and generated images (reflected in their competitive SSIM scores) their outputs tend to be over-smoothed, reducing the perceptual realism of the generated images and lowering PSNR. \textbf{Image translation} represents the most challenging task evaluated in this work, apart from the DDPM, all the baseline methods struggled to produce high-quality generations. NOIR outperforms all the baselines on the lowest resolution, achieves performance comparable to ViT on the high resolutions, and maintains its performance across all tested resolutions.

Section~\ref{eps-ReNO-experiments} empirically demonstrates a key theoretical contribution of this work: \textbf{NOIR is an $\epsilon$-ReNO} in the sense of Bartolucci \textit{et al.}~\cite{Bartolucci2023-hv}, a desirable property for NOs that FNO, for instance, does not satisfy. As shown in Figure~\ref{fig:dice_distribution}, the distribution of DSCs at resolutions $32^2$, $64^2$, $128^2$, $160^2$, and $200^2$ remains highly consistent, with mean Dice ranging from $0.93$ to $0.94$, confirming minimal aliasing error under large resolution changes. NOIR achieves this property by operating in function space. The input INR requires only sparse observations to recover the underlying continuous function, yielding a latent modulation that is independent of the discretization. Since the NO acts solely on this latent representation, it produces the same predicted modulation regardless of the input resolution. The output INR can then be evaluated at any desired resolution. Another key property of NOIR is its modular and task-agnostic design. Since the input INR, output INR, and NO are trained separately rather than end-to-end, the learned INRs can be reused across different downstream tasks, requiring only the NO and one of the two INRs to be adapted for a new task.

Despite these promising results, NOIR has some \textbf{limitations} that warrant discussion. The performance of the framework depends critically on the ability of INRs to accurately approximate continuous functions at the dataset level. A central factor in this approximation is the dimensionality of the latent representation, as it constitutes the sole carrier of signal-specific information. For complex images with high inter-patient variability and rich fine-grained detail, this representation must be sufficiently large to capture the relevant signal content. However, as the latent dimensionality grows, training the NO that maps between representations becomes substantially more challenging, requiring larger datasets and more expressive architectures as the number of learnable parameters scales accordingly. The choice of operator architecture is therefore not independent of the latent space dimensionality, and careful co-design of both components is necessary to achieve reliable performance across tasks. A further limitation of NOIR in its current form is its reliance on supervised learning. Training the framework requires paired data, as both the input and output signals must be independently encoded into their respective INRs before training the NO. This requirement limits its use to tasks with supervision or annotated target data, thereby restricting its applicability when such annotations are unavailable. For instance, in image registration, ground-truth displacement fields are typically not available and learning is performed in an unsupervised manner. Extending NOIR to unsupervised or weakly supervised settings is an important direction for future work.

\section{Conclusion}
We introduced NOIR, a framework to challenge mainstream grid-based deep learning approaches not only conceptually but also empirically. By combining INRs with operator learning, the framework provides competitive performance while introducing additional advantages in resolution robustness, task flexibility, and architectural simplicity. Future work will investigate the trade-off between modulation size and training stability of the NO, as larger latent representations improve signal fidelity but increase the complexity of the operator learning problem. Furthermore, we will focus on extending NOIR to unsupervised learning of downstream medical imaging tasks, in particular, deformable image registration.

\bibliographystyle{splncs04}
\bibliography{main}

@INPROCEEDINGS{Xiang2025-hy,
  title           = "A comprehensive review on applications of computer vision
                     in medical imaging",
  booktitle       = "Fifth International Conference on Computer Vision and
                     Pattern Analysis ({ICCPA} 2025)",
  author          = "Xiang, Bangyin",
  editor          = "Chen, Chunyi and Lei, Tao",
  publisher       = "SPIE",
  pages           = "11",
  month           =  aug,
  year            =  2025,
  conference      = "Fifth International Conference on Computer Vision and
                     Pattern Analysis (ICCPA 2025)",
  location        = "Anshan, China"
}

@ARTICLE{Liu2021-jv,
  title     = "Advances in deep learning-based medical image analysis",
  author    = "Liu, Xiaoqing and Gao, Kunlun and Liu, Bo and Pan, Chengwei and
               Liang, Kongming and Yan, Lifeng and Ma, Jiechao and He, Fujin
               and Zhang, Shu and Pan, Siyuan and Yu, Yizhou",
  journal   = "Health Data Sci.",
  publisher = "American Association for the Advancement of Science (AAAS)",
  volume    =  2021,
  pages     = "8786793",
  month     =  may,
  year      =  2021,
  copyright = "http://creativecommons.org/licenses/by/4.0/",
  language  = "en"
}

@ARTICLE{Gao2025-jf,
  title     = "Medical image segmentation: A comprehensive review of deep
               learning-based methods",
  author    = "Gao, Yuxiao and Jiang, Yang and Peng, Yanhong and Yuan, Fujiang
               and Zhang, Xinyue and Wang, Jianfeng",
  journal   = "Tomography",
  publisher = "MDPI AG",
  volume    =  11,
  number    =  5,
  pages     = "52",
  month     =  apr,
  year      =  2025,
  copyright = "https://creativecommons.org/licenses/by/4.0/",
  language  = "en"
}

@ARTICLE{Darzi2024-zp,
  title     = "A review of medical image registration for different modalities",
  author    = "Darzi, Fatemehzahra and Bocklitz, Thomas",
  journal   = "Bioengineering (Basel)",
  publisher = "MDPI AG",
  volume    =  11,
  number    =  8,
  pages     = "786",
  month     =  aug,
  year      =  2024,
  copyright = "https://creativecommons.org/licenses/by/4.0/",
  language  = "en"
}

@ARTICLE{Chen2025-cw,
  title     = "Medical image translation with deep learning: Advances, datasets
               and perspectives",
  author    = "Chen, Junxin and Ye, Zhiheng and Zhang, Renlong and Li, Hao and
               Fang, Bo and Zhang, Li-Bo and Wang, Wei",
  journal   = "Med. Image Anal.",
  publisher = "Elsevier BV",
  volume    =  103,
  number    =  103605,
  pages     = "103605",
  month     =  jul,
  year      =  2025,
  language  = "en"
}

@ARTICLE{Suganyadevi2022-ht,
  title     = "A review on deep learning in medical image analysis",
  author    = "Suganyadevi, S and Seethalakshmi, V and Balasamy, K",
  journal   = "Int. J. Multimed. Inf. Retr.",
  publisher = "Springer Science and Business Media LLC",
  volume    =  11,
  number    =  1,
  pages     = "19--38",
  year      =  2022,
  copyright = "https://www.springernature.com/gp/researchers/text-and-data-mining",
  language  = "en"
}

@INPROCEEDINGS{Molaei2023-aa,
  title           = "Implicit neural representation in medical imaging: A
                     comparative survey",
  booktitle       = "2023 {IEEE/CVF} International Conference on Computer
                     Vision Workshops ({ICCVW})",
  author          = "Molaei, Amirali and Aminimehr, Amirhossein and Tavakoli,
                     Armin and Kazerouni, Amirhossein and Azad, Bobby and Azad,
                     Reza and Merhof, Dorit",
  publisher       = "IEEE",
  pages           = "2373--2383",
  month           =  oct,
  year            =  2023,
  conference      = "2023 IEEE/CVF International Conference on Computer Vision
                     Workshops (ICCVW)",
  location        = "Paris, France"
}

@ARTICLE{Sitzmann2020-yx,
  title        = "Implicit neural representations with periodic activation
                  functions",
  author       = "Sitzmann, Vincent and Martel, Julien N P and Bergman,
                  Alexander W and Lindell, David B and Wetzstein, Gordon",
  year         =  2020,
  primaryClass = "cs.CV",
  eprint       = "2006.09661"
}

@ARTICLE{Genova2019-nx,
  title         = "Learning shape templates with structured implicit functions",
  author        = "Genova, Kyle and Cole, Forrester and Vlasic, Daniel and
                   Sarna, Aaron and Freeman, William T and Funkhouser, Thomas",
  month         =  apr,
  year          =  2019,
  copyright     = "http://arxiv.org/licenses/nonexclusive-distrib/1.0/",
  archivePrefix = "arXiv",
  primaryClass  = "cs.CV",
  eprint        = "1904.06447"
}

@ARTICLE{Tancik2020-xl,
  title         = "Fourier features let networks learn high frequency functions
                   in low dimensional domains",
  author        = "Tancik, Matthew and Srinivasan, Pratul P and Mildenhall, Ben
                   and Fridovich-Keil, Sara and Raghavan, Nithin and Singhal,
                   Utkarsh and Ramamoorthi, Ravi and Barron, Jonathan T and Ng,
                   Ren",
  month         =  jun,
  year          =  2020,
  copyright     = "http://arxiv.org/licenses/nonexclusive-distrib/1.0/",
  archivePrefix = "arXiv",
  primaryClass  = "cs.CV",
  eprint        = "2006.10739"
}

@ARTICLE{Muller2022-mf,
  title         = "Instant neural graphics primitives with a multiresolution
                   hash encoding",
  author        = "M{\"u}ller, Thomas and Evans, Alex and Schied, Christoph and
                   Keller, Alexander",
  month         =  jan,
  year          =  2022,
  copyright     = "http://arxiv.org/licenses/nonexclusive-distrib/1.0/",
  archivePrefix = "arXiv",
  primaryClass  = "cs.CV",
  eprint        = "2201.05989"
}

@ARTICLE{Mildenhall2020-ng,
  title         = "{NeRF}: Representing scenes as neural radiance fields for
                   view synthesis",
  author        = "Mildenhall, Ben and Srinivasan, Pratul P and Tancik, Matthew
                   and Barron, Jonathan T and Ramamoorthi, Ravi and Ng, Ren",
  month         =  mar,
  year          =  2020,
  copyright     = "http://arxiv.org/licenses/nonexclusive-distrib/1.0/",
  archivePrefix = "arXiv",
  primaryClass  = "cs.CV",
  eprint        = "2003.08934"
}

@ARTICLE{Fan2024-zm,
  title         = "Optimizing {3D} geometry reconstruction from implicit neural
                   representations",
  author        = "Fan, Shen and Musialski, Przemyslaw",
  month         =  oct,
  year          =  2024,
  copyright     = "http://creativecommons.org/licenses/by-nc-nd/4.0/",
  archivePrefix = "arXiv",
  primaryClass  = "cs.CV",
  eprint        = "2410.12725"
}

@ARTICLE{Dupont2022-qm,
  title         = "{COIN++}: Neural Compression Across Modalities",
  author        = "Dupont, Emilien and Loya, Hrushikesh and Alizadeh, Milad and
                   Goli{\'n}ski, Adam and Teh, Yee Whye and Doucet, Arnaud",
  month         =  jan,
  year          =  2022,
  copyright     = "http://arxiv.org/licenses/nonexclusive-distrib/1.0/",
  archivePrefix = "arXiv",
  primaryClass  = "cs.LG",
  eprint        = "2201.12904"
}

@ARTICLE{Ha2016-vc,
  title         = "{HyperNetworks}",
  author        = "Ha, David and Dai, Andrew and Le, Quoc V",
  month         =  sep,
  year          =  2016,
  copyright     = "http://arxiv.org/licenses/nonexclusive-distrib/1.0/",
  archivePrefix = "arXiv",
  primaryClass  = "cs.LG",
  eprint        = "1609.09106"
}

@ARTICLE{Sitzmann2020-dy,
  title         = "{MetaSDF}: Meta-learning signed distance functions",
  author        = "Sitzmann, Vincent and Chan, Eric R and Tucker, Richard and
                   Snavely, Noah and Wetzstein, Gordon",
  month         =  jun,
  year          =  2020,
  copyright     = "http://arxiv.org/licenses/nonexclusive-distrib/1.0/",
  archivePrefix = "arXiv",
  primaryClass  = "cs.CV",
  eprint        = "2006.09662"
}

@ARTICLE{Park2019-kp,
  title         = "Semantic image synthesis with spatially-adaptive
                   normalization",
  author        = "Park, Taesung and Liu, Ming-Yu and Wang, Ting-Chun and Zhu,
                   Jun-Yan",
  month         =  mar,
  year          =  2019,
  copyright     = "http://arxiv.org/licenses/nonexclusive-distrib/1.0/",
  archivePrefix = "arXiv",
  primaryClass  = "cs.CV",
  eprint        = "1903.07291"
}

@ARTICLE{Dupont2022-kn,
  title         = "From data to functa: Your data point is a function and you
                   can treat it like one",
  author        = "Dupont, Emilien and Kim, Hyunjik and Eslami, S M Ali and
                   Rezende, Danilo and Rosenbaum, Dan",
  month         =  jan,
  year          =  2022,
  copyright     = "http://arxiv.org/licenses/nonexclusive-distrib/1.0/",
  archivePrefix = "arXiv",
  primaryClass  = "cs.LG",
  eprint        = "2201.12204"
}

@ARTICLE{Kazerouni2025-lb,
  title         = "{LIFT}: Latent implicit functions for task- and
                   data-agnostic encoding",
  author        = "Kazerouni, Amirhossein and Mehraban, Soroush and Brudno,
                   Michael and Taati, Babak",
  month         =  mar,
  year          =  2025,
  copyright     = "http://creativecommons.org/licenses/by/4.0/",
  archivePrefix = "arXiv",
  primaryClass  = "cs.LG",
  eprint        = "2503.15420"
}

@article{friedrich2025medfuncta,
  title={MedFuncta: A Unified Framework for Learning Efficient Medical Neural Fields},
  author={Friedrich, Paul and Bieder, Florentin and McGinnis, Julian and Wolleb, Julia and Rueckert, Daniel and Cattin, Philippe C},
  journal={arXiv preprint arXiv:2502.14401},
  year={2025}
}

@ARTICLE{Kovachki2021-ch,
  title         = "Neural operator: Learning maps between function spaces",
  author        = "Kovachki, Nikola and Li, Zongyi and Liu, Burigede and
                   Azizzadenesheli, Kamyar and Bhattacharya, Kaushik and
                   Stuart, Andrew and Anandkumar, Anima",
  month         =  aug,
  year          =  2021,
  copyright     = "http://arxiv.org/licenses/nonexclusive-distrib/1.0/",
  archivePrefix = "arXiv",
  primaryClass  = "cs.LG",
  eprint        = "2108.08481"
}

@ARTICLE{Lu2019-tv,
  title         = "{DeepONet}: Learning nonlinear operators for identifying
                   differential equations based on the universal approximation
                   theorem of operators",
  author        = "Lu, Lu and Jin, Pengzhan and Karniadakis, George Em",
  month         =  oct,
  year          =  2019,
  copyright     = "http://creativecommons.org/licenses/by-nc-sa/4.0/",
  archivePrefix = "arXiv",
  primaryClass  = "cs.LG",
  eprint        = "1910.03193"
}

@ARTICLE{Li2020-ss,
  title         = "Fourier neural operator for parametric partial differential
                   equations",
  author        = "Li, Zongyi and Kovachki, Nikola and Azizzadenesheli, Kamyar
                   and Liu, Burigede and Bhattacharya, Kaushik and Stuart,
                   Andrew and Anandkumar, Anima",
  month         =  oct,
  year          =  2020,
  copyright     = "http://arxiv.org/licenses/nonexclusive-distrib/1.0/",
  archivePrefix = "arXiv",
  primaryClass  = "cs.LG",
  eprint        = "2010.08895"
}

@MISC{Raonic_undated-gw,
  title        = "Convolutional Neural Operators for robust and accurate
                  learning of {PDEs}",
  author       = "Raoni{\'c}, Bogdan and Molinaro, Roberto and De Ryck, Tim and
                  Rohner, Tobias and Bartolucci, Francesca and Alaifari, Rima
                  and Mishra, Siddhartha and De B{\'e}zenac, Emmanuel",
  howpublished = "\url{https://proceedings.neurips.cc/paper_files/paper/2023/file/f3c1951b34f7f55ffaecada7fde6bd5a-Paper-Conference.pdf}",
  note         = "Accessed: 2026-2-1"
}

@ARTICLE{Li2020-nb,
  title         = "Neural operator: Graph kernel network for partial
                   differential equations",
  author        = "Li, Zongyi and Kovachki, Nikola and Azizzadenesheli, Kamyar
                   and Liu, Burigede and Bhattacharya, Kaushik and Stuart,
                   Andrew and Anandkumar, Anima",
  month         =  mar,
  year          =  2020,
  copyright     = "http://arxiv.org/licenses/nonexclusive-distrib/1.0/",
  archivePrefix = "arXiv",
  primaryClass  = "cs.LG",
  eprint        = "2003.03485"
}

@ARTICLE{Serrano2023-jl,
  title         = "Operator learning with neural fields: Tackling {PDEs} on
                   general geometries",
  author        = "Serrano, Louis and Boudec, Lise Le and Koupa{\"\i}, Armand
                   Kassa{\"\i} and Wang, Thomas X and Yin, Yuan and Vittaut,
                   Jean-No{\"e}l and Gallinari, Patrick",
  month         =  jun,
  year          =  2023,
  copyright     = "http://arxiv.org/licenses/nonexclusive-distrib/1.0/",
  archivePrefix = "arXiv",
  primaryClass  = "cs.LG",
  eprint        = "2306.07266"
}

@ARTICLE{Ronneberger2015-ql,
  title         = "{U-Net}: Convolutional Networks for Biomedical Image
                   Segmentation",
  author        = "Ronneberger, Olaf and Fischer, Philipp and Brox, Thomas",
  month         =  may,
  year          =  2015,
  copyright     = "http://arxiv.org/licenses/nonexclusive-distrib/1.0/",
  archivePrefix = "arXiv",
  primaryClass  = "cs.CV",
  eprint        = "1505.04597"
}

@ARTICLE{Fabian2018-ay,
  title         = "{NnU-Net}: Self-adapting framework for {U-Net-based} medical
                   image segmentation",
  author        = "Fabian, Isensee and Jens, Petersen and Andre, Klein and
                   David, Zimmerer and Paul, F Jaeger and Simon, Kohl and
                   Jakob, Wasserthal and Gregor, Koehler and Tobias, Norajitra
                   and Sebastian, Wirkert and Klaus, H Maier-Hein",
  month         =  sep,
  year          =  2018,
  archivePrefix = "arXiv",
  primaryClass  = "cs.CV",
  eprint        = "1809.10486"
}

@ARTICLE{Chen2021-ew,
  title         = "{TransUNet}: Transformers make strong encoders for medical
                   image segmentation",
  author        = "Chen, Jieneng and Lu, Yongyi and Yu, Qihang and Luo, Xiangde
                   and Adeli, Ehsan and Wang, Yan and Lu, Le and Yuille, Alan L
                   and Zhou, Yuyin",
  month         =  feb,
  year          =  2021,
  copyright     = "http://creativecommons.org/licenses/by/4.0/",
  archivePrefix = "arXiv",
  primaryClass  = "cs.CV",
  eprint        = "2102.04306"
}

@ARTICLE{Dosovitskiy2020-ig,
  title         = "An image is worth 16x16 words: Transformers for image
                   recognition at scale",
  author        = "Dosovitskiy, Alexey and Beyer, Lucas and Kolesnikov,
                   Alexander and Weissenborn, Dirk and Zhai, Xiaohua and
                   Unterthiner, Thomas and Dehghani, Mostafa and Minderer,
                   Matthias and Heigold, Georg and Gelly, Sylvain and
                   Uszkoreit, Jakob and Houlsby, Neil",
  month         =  oct,
  year          =  2020,
  copyright     = "http://arxiv.org/licenses/nonexclusive-distrib/1.0/",
  archivePrefix = "arXiv",
  primaryClass  = "cs.CV",
  eprint        = "2010.11929"
}

@ARTICLE{Hu2021-et,
  title         = "{Swin-Unet}: Unet-like Pure Transformer for Medical Image
                   Segmentation",
  author        = "Hu, Cao and Yueyue, Wang and Joy, Chen and Dongsheng, Jiang
                   and Xiaopeng, Zhang and Qi, Tian and Manning, Wang",
  month         =  may,
  year          =  2021,
  archivePrefix = "arXiv",
  primaryClass  = "eess.IV",
  eprint        = "2105.05537"
}

@ARTICLE{Cai2023-xb,
  title     = "{CycleGAN-based} image translation from {MRI} to {CT} scans",
  author    = "Cai, Yingchao and Li, Mengxiao and Liu, Shiqi and Zhou, Changhao",
  journal   = "J. Phys. Conf. Ser.",
  publisher = "IOP Publishing",
  volume    =  2646,
  number    =  1,
  pages     = "012016",
  month     =  dec,
  year      =  2023,
  copyright = "http://creativecommons.org/licenses/by/3.0/"
}

@ARTICLE{Baldini2023-oy,
  title         = "{MRI} scan synthesis methods based on clustering and
                   {Pix2Pix}",
  author        = "Baldini, Giulia and Schmidt, Melanie and Z{\"a}ske,
                   Charlotte and Caldeira, Liliana L",
  month         =  dec,
  year          =  2023,
  copyright     = "http://arxiv.org/licenses/nonexclusive-distrib/1.0/",
  archivePrefix = "arXiv",
  primaryClass  = "eess.IV",
  eprint        = "2312.05176"
}

@ARTICLE{Ho2020-dq,
  title         = "Denoising Diffusion Probabilistic Models",
  author        = "Ho, Jonathan and Jain, Ajay and Abbeel, Pieter",
  month         =  jun,
  year          =  2020,
  copyright     = "http://arxiv.org/licenses/nonexclusive-distrib/1.0/",
  archivePrefix = "arXiv",
  primaryClass  = "cs.LG",
  eprint        = "2006.11239"
}

@ARTICLE{Friedrich2024-pu,
  title         = "{WDM}: {3D} wavelet Diffusion Models for high-resolution
                   medical image synthesis",
  author        = "Friedrich, Paul and Wolleb, Julia and Bieder, Florentin and
                   Durrer, Alicia and Cattin, Philippe C",
  month         =  feb,
  year          =  2024,
  copyright     = "http://arxiv.org/licenses/nonexclusive-distrib/1.0/",
  archivePrefix = "arXiv",
  primaryClass  = "eess.IV",
  eprint        = "2402.19043"
}

@ARTICLE{Durrer2025-id,
  title         = "{fastWDM3D}: Fast and Accurate {3D} Healthy Tissue
                   Inpainting",
  author        = "Durrer, Alicia and Bieder, Florentin and Friedrich, Paul and
                   Menze, Bjoern and Cattin, Philippe C and Kofler, Florian",
  month         =  jul,
  year          =  2025,
  copyright     = "http://arxiv.org/licenses/nonexclusive-distrib/1.0/",
  archivePrefix = "arXiv",
  primaryClass  = "eess.IV",
  eprint        = "2507.13146"
}

@ARTICLE{Balakrishnan2018-ef,
  title         = "{VoxelMorph}: A learning framework for deformable medical
                   image registration",
  author        = "Balakrishnan, Guha and Zhao, Amy and Sabuncu, Mert R and
                   Guttag, John and Dalca, Adrian V",
  month         =  sep,
  year          =  2018,
  copyright     = "http://arxiv.org/licenses/nonexclusive-distrib/1.0/",
  archivePrefix = "arXiv",
  primaryClass  = "cs.CV",
  eprint        = "1809.05231"
}

@ARTICLE{Wong2023-gk,
  title         = "{FNOSeg3D}: Resolution-robust {3D} image segmentation with
                   Fourier neural operator",
  author        = "Wong, Ken C L and Hongzhi, Wang and Tanveer, Syeda-Mahmood",
  month         =  oct,
  year          =  2023,
  archivePrefix = "arXiv",
  primaryClass  = "eess.IV",
  eprint        = "2310.03872"
}

@INCOLLECTION{Drozdov2025-br,
  title     = "{FNOReg}: Resolution-robust medical image registration method
               based on Fourier neural operator",
  booktitle = "Lecture Notes in Computer Science",
  author    = "Drozdov, Nikita A and Sorokin, Dmitry V",
  publisher = "Springer Nature Switzerland",
  pages     = "163--177",
  series    = "Lecture Notes in Computer Science",
  year      =  2025,
  address   = "Cham",
  copyright = "https://www.springernature.com/gp/researchers/text-and-data-mining",
  language  = "en"
}

@ARTICLE{Jaeger2014-kh,
  title    = "Two public chest X-ray datasets for computer-aided screening of
              pulmonary diseases",
  author   = "Jaeger, Stefan and Candemir, Sema and Antani, Sameer and
              W{\'a}ng, Y{\`\i}-Xi{\'a}ng J and Lu, Pu-Xuan and Thoma, George",
  journal  = "Quant. Imaging Med. Surg.",
  volume   =  4,
  number   =  6,
  pages    = "475--477",
  month    =  dec,
  year     =  2014,
  language = "en"
}

@ARTICLE{Marcus2007-or,
  title     = "Open Access Series of Imaging Studies ({OASIS)}: cross-sectional
               {MRI} data in young, middle aged, nondemented, and demented
               older adults",
  author    = "Marcus, Daniel S and Wang, Tracy H and Parker, Jamie and
               Csernansky, John G and Morris, John C and Buckner, Randy L",
  journal   = "J. Cogn. Neurosci.",
  publisher = "MIT Press - Journals",
  volume    =  19,
  number    =  9,
  pages     = "1498--1507",
  month     =  sep,
  year      =  2007,
  language  = "en"
}

@ARTICLE{Kodym2021-xv,
  title     = "{SkullBreak} / {SkullFix} - Dataset for automatic cranial
               implant design and a benchmark for volumetric shape learning
               tasks",
  author    = "Kodym, Old{\v r}ich and Li, Jianning and Pepe, Antonio and
               Gsaxner, Christina and Chilamkurthy, Sasank and Egger, Jan and
               {\v S}pan{\v e}l, Michal",
  journal   = "Data Brief",
  publisher = "Elsevier BV",
  volume    =  35,
  number    =  106902,
  pages     = "106902",
  month     =  apr,
  year      =  2021,
  copyright = "http://creativecommons.org/licenses/by-nc-nd/4.0/",
  language  = "en"
}

@ARTICLE{Knoll2020-iu,
  title     = "{FastMRI}: A publicly available raw k-space and {DICOM} dataset
               of knee images for accelerated {MR} image reconstruction using
               machine learning",
  author    = "Knoll, Florian and Zbontar, Jure and Sriram, Anuroop and
               Muckley, Matthew J and Bruno, Mary and Defazio, Aaron and
               Parente, Marc and Geras, Krzysztof J and Katsnelson, Joe and
               Chandarana, Hersh and Zhang, Zizhao and Drozdzalv, Michal and
               Romero, Adriana and Rabbat, Michael and Vincent, Pascal and
               Pinkerton, James and Wang, Duo and Yakubova, Nafissa and Owens,
               Erich and Zitnick, C Lawrence and Recht, Michael P and
               Sodickson, Daniel K and Lui, Yvonne W",
  journal   = "Radiol. Artif. Intell.",
  publisher = "Radiological Society of North America (RSNA)",
  volume    =  2,
  number    =  1,
  pages     = "e190007",
  month     =  jan,
  year      =  2020,
  language  = "en"
}

@inproceedings{friedrich2023point,
  title={Point cloud diffusion models for automatic implant generation},
  author={Friedrich, Paul and Wolleb, Julia and Bieder, Florentin and Thieringer, Florian M and Cattin, Philippe C},
  booktitle={International conference on medical image computing and computer-assisted intervention},
  pages={112--122},
  year={2023},
  organization={Springer}
}

@inproceedings{tian2023gradicon,
  title={Gradicon: Approximate diffeomorphisms via gradient inverse consistency},
  author={Tian, Lin and Greer, Hastings and Vialard, Fran{\c{c}}ois-Xavier and Kwitt, Roland and Est{\'e}par, Ra{\'u}l San Jos{\'e} and Rushmore, Richard Jarrett and Makris, Nikolaos and Bouix, Sylvain and Niethammer, Marc},
  booktitle={Proceedings of the IEEE/CVF Conference on Computer Vision and Pattern Recognition},
  pages={18084--18094},
  year={2023}
}

@ARTICLE{Oktay2018-hr,
  title        = "Attention {U-Net}: Learning where to look for the pancreas",
  author       = "Oktay, Ozan and Schlemper, Jo and Folgoc, Loic Le and Lee,
                  Matthew and Heinrich, Mattias and Misawa, Kazunari and Mori,
                  Kensaku and McDonagh, Steven and Hammerla, Nils Y and Kainz,
                  Bernhard and Glocker, Ben and Rueckert, Daniel",
  year         =  2018,
  primaryClass = "cs.CV",
  eprint       = "1804.03999"
}

@inproceedings{gafencu2025shape,
  title={Shape Completion and Real-Time Visualization in Robotic Ultrasound Spine Acquisitions},
  author={Gafencu, Miruna-Alexandra and Shaban, Reem and Velikova, Yordanka and Azampour, Mohammad Farid and Navab, Nassir},
  booktitle={2025 IEEE/RSJ International Conference on Intelligent Robots and Systems (IROS)},
  pages={20906--20911},
  year={2025},
  organization={IEEE}
}

@article{schlemper2017deep,
  title={A deep cascade of convolutional neural networks for dynamic MR image reconstruction},
  author={Schlemper, Jo and Caballero, Jose and Hajnal, Joseph V and Price, Anthony N and Rueckert, Daniel},
  journal={IEEE transactions on Medical Imaging},
  volume={37},
  number={2},
  pages={491--503},
  year={2017},
  publisher={IEEE}
}

@inproceedings{li2024classification,
  title={Classification, Regression and Segmentation directly from k-Space in Cardiac MRI},
  author={Li, Ruochen and Pan, Jiazhen and Zhu, Youxiang and Ni, Juncheng and Rueckert, Daniel},
  booktitle={International Workshop on Machine Learning in Medical Imaging},
  pages={31--41},
  year={2024},
  organization={Springer}
}

@article{li2024novel,
  title={A novel automatic segmentation method directly based on magnetic resonance imaging K-space data for auxiliary diagnosis of glioma},
  author={Li, Yikang and Qi, Yulong and Hu, Zhanli and Zhang, Ke and Jia, Sen and Zhang, Lei and Xu, Wenjing and Shen, Shuai and W{\'a}ng, Y{\`\i} Xi{\'a}ng J and Li, Zongyang and others},
  journal={Quantitative Imaging in Medicine and Surgery},
  volume={14},
  number={2},
  pages={2008--2020},
  year={2024},
  publisher={LWW}
}

@article{rempe2024k,
  title={k-strip: A novel segmentation algorithm in k-space for the application of skull stripping},
  author={Rempe, Moritz and Mentzel, Florian and Pomykala, Kelsey L and Haubold, Johannes and Nensa, Felix and Kroeninger, Kevin and Egger, Jan and Kleesiek, Jens},
  journal={Computer Methods and Programs in Biomedicine},
  volume={243},
  pages={107912},
  year={2024},
  publisher={Elsevier}
}

@inproceedings{litany2017deep,
  title={Deep functional maps: Structured prediction for dense shape correspondence},
  author={Litany, Or and Remez, Tal and Rodola, Emanuele and Bronstein, Alex and Bronstein, Michael},
  booktitle={Proceedings of the IEEE international conference on computer vision},
  pages={5659--5667},
  year={2017}
}

@inproceedings{li2025fm,
  title={FM-POT: non-rigid point cloud registration based on functional maps and partial optimal transport fusion},
  author={Li, Siyu and Li, Minqi},
  booktitle={Fifth International Conference on Optical Imaging and Image Processing (ICOIP 2025)},
  volume={13688},
  pages={389--396},
  year={2025},
  organization={SPIE}
}

@article{Zhu2018_AUTOMAP,
author = {Zhu, Bo and Liu, Jeremiah Z. and Cauley, Stephen F. and Rosen, Bruce R. and Rosen, Mikhail S.},
title = {Image reconstruction by domain-transform manifold learning},
journal = {Nature},
volume = {555},
pages = {487--492},
year = {2018},
doi = {10.1038/nature25988}
}

@article{khan2020adaptive,
  title={Adaptive and compressive beamforming using deep learning for medical ultrasound},
  author={Khan, Shujaat and Huh, Jaeyoung and Ye, Jong Chul},
  journal={IEEE transactions on ultrasonics, ferroelectrics, and frequency control},
  volume={67},
  number={8},
  pages={1558--1572},
  year={2020},
  publisher={IEEE}
}

@article{han2018framing,
  title={Framing U-Net via deep convolutional framelets: Application to sparse-view CT},
  author={Han, Yoseob and Ye, Jong Chul},
  journal={IEEE transactions on medical imaging},
  volume={37},
  number={6},
  pages={1418--1429},
  year={2018},
  publisher={IEEE}
}

@ARTICLE{Bartolucci2023-hv,
  title         = "Representation equivalent Neural Operators: A framework for
                   alias-free operator learning",
  author        = "Bartolucci, Francesca and de B{\'e}zenac, Emmanuel and
                   Raoni{\'c}, Bogdan and Molinaro, Roberto and Mishra,
                   Siddhartha and Alaifari, Rima",
  month         =  may,
  year          =  2023,
  copyright     = "http://arxiv.org/licenses/nonexclusive-distrib/1.0/",
  archivePrefix = "arXiv",
  primaryClass  = "cs.LG",
  eprint        = "2305.19913"
}

@ARTICLE{Rempe2024-ks,
  title     = "k-strip: A novel segmentation algorithm in k-space for the
               application of skull stripping",
  author    = "Rempe, Moritz and Mentzel, Florian and Pomykala, Kelsey L and
               Haubold, Johannes and Nensa, Felix and Kroeninger, Kevin and
               Egger, Jan and Kleesiek, Jens",
  journal   = "Comput. Methods Programs Biomed.",
  publisher = "Elsevier BV",
  volume    =  243,
  number    =  107912,
  pages     = "107912",
  month     =  jan,
  year      =  2024,
  copyright = "http://creativecommons.org/licenses/by-nc-nd/4.0/",
  language  = "en"
}

@article{friedrich2024cwdm,
                        title={cWDM: Conditional Wavelet Diffusion Models for Cross-Modality 3D Medical Image Synthesis},
                        author={Friedrich, Paul and Durrer, Alicia and Wolleb, Julia and Cattin, Philippe C},
                        journal={arXiv preprint arXiv:2411.17203},
                        year={2024}}

@INCOLLECTION{El_hadramy2023-la,
  title     = "Intraoperative {CT} augmentation for needle-based liver
               interventions",
  booktitle = "Lecture Notes in Computer Science",
  author    = "El hadramy, Sidaty and Verde, Juan and Padoy, Nicolas and Cotin,
               St{\'e}phane",
  publisher = "Springer Nature Switzerland",
  pages     = "291--301",
  series    = "Lecture Notes in Computer Science",
  year      =  2023,
  address   = "Cham",
  copyright = "https://www.springernature.com/gp/researchers/text-and-data-mining",
  language  = "en"
}

@ARTICLE{Vyas2025-ez,
  title         = "Fit pixels, get labels: Meta-learned implicit networks for
                   image segmentation",
  author        = "Vyas, Kushal and Veeraraghavan, Ashok and Balakrishnan, Guha",
  month         =  oct,
  year          =  2025,
  copyright     = "http://creativecommons.org/licenses/by/4.0/",
  archivePrefix = "arXiv",
  primaryClass  = "cs.CV",
  eprint        = "2510.04021"
}

@INCOLLECTION{Hadramy2026-gb,
  title     = "cIDIR: Conditioned implicit
               neural representation for regularized deformable image
               registration",
  booktitle = "Lecture Notes in Computer Science",
  author    = "Hadramy, Sidaty El and Cherkaoui, Oumeymah and Cattin, Philippe
               C",
  publisher = "Springer Nature Switzerland",
  pages     = "87--96",
  series    = "Lecture Notes in Computer Science",
  year      =  2026,
  address   = "Cham",
  copyright = "https://www.springernature.com/gp/researchers/text-and-data-mining",
  language  = "en"
}

% ================================================================
% SUPPLEMENTARY MATERIAL
% ================================================================

\newpage

% Reset figure, table, and equation counters with S prefix
\setcounter{figure}{0}
\setcounter{table}{0}
\setcounter{equation}{0}
\renewcommand{\thefigure}{S\arabic{figure}}
\renewcommand{\thetable}{S\arabic{table}}
\renewcommand{\theequation}{S\arabic{equation}}

\begin{center}

{\Large \textbf{Supplementary Material}}

\vspace{0.6cm}

\end{center}

In this supplementary material we provide additional implementation details,
reproducibility information, and extended experimental results that complement
the main paper. In particular, we report detailed training configurations,
dataset splits, additional qualitative reconstructions, comparisons with
baseline models, and further analysis of the proposed method.

\section*{Reproducibility Details}

To ensure reproducibility of our experiments, we summarize all architectures
and training hyperparameters used across datasets in the tables below. Table~\ref{tab:input_inr} reports the training configuration for the input implicit neural representations (INRs), Table~\ref{tab:output_inr} reports the configuration for the output INRs, and Table~\ref{tab:neural_operator} reports the configuration for the neural operator (NO) that maps between the input and output INRs. For clarity, we adopt the following abbreviations for column names: \textbf{In} denotes the input size, \textbf{Out} the output size, \textbf{\#HL} the number of hidden layers, \textbf{HS} the hidden layer size, \textbf{HAct} the hidden layer activation function, \textbf{LAct} the activation function of the last layer, and \textbf{Lat} the size of the latent representation. Additionally, \textbf{\#HHL} and \textbf{HHS} denote the number of hidden layers and the size of hidden layers in the hypernetwork, respectively. \textbf{LRL} denotes the learning rate for the latent representation, \textbf{LR} the learning rate, \textbf{Opt} the optimizer, \textbf{WD} the weight decay, and \textbf{Dropout} the dropout rate. For the neural operator tables, \textbf{HD} refers to the hidden dimension of each layer, \textbf{Act} to the activation function, and \textbf{Res} indicates whether residual connections are used. The datasets used, their image sizes, splits, and the number of sampled points are summarized in Table~\ref{tab:datasets}.

% ------------------------------------------------
\begin{table*}[h]
\centering
\caption{Training configuration for input implicit neural representations (INRs).}
\label{tab:input_inr}
\resizebox{\textwidth}{!}{
\begin{tabular}{lcccccccccccc}
\toprule
Dataset  & In & Out & \#HL & HS & HAct & LAct & Lat & \#HHL & HHS & LRL & LR & Opt \\
\midrule
Shenzhen   & 2 & 1 & 6 & 256 & \texttt{Sine} & \texttt{Sigmoid} & 64 & 1 & 64 & $10^{-2}$ & $5\times10^{-6}$ & \texttt{AdamW} \\
OASIS-4    & 2 & 1 & 6 & 256 & \texttt{Sine} & \texttt{Sigmoid} & 2048 & 1 & 64 & $10^{-2}$ & $5\times10^{-6}$ & \texttt{AdamW} \\
SkullBreak & 3 & 1 & 10 & 256 & \texttt{Sine} & \texttt{Softmax} & 64 & 1 & 64 & $10^{-2}$ & $5\times10^{-6}$ & \texttt{AdamW} \\
fastMRI    & 2 & 1 & 10 & 256 & \texttt{Sine} & \texttt{Sigmoid} & 4096 & 1 & 64 & $10^{-2}$ & $5\times10^{-6}$ & \texttt{AdamW} \\
Ultrasound & 2 & 1 & 10 & 256 & \texttt{Sine} & \texttt{Sigmoid} & 64 & 1 & 64 & $10^{-2}$ & $5\times10^{-6}$ & \texttt{AdamW} \\
\bottomrule
\end{tabular}
}
\end{table*}

% ------------------------------------------------
\begin{table*}[h]
\centering
\caption{Training configuration for output implicit neural representations (INRs).}
\label{tab:output_inr}
\resizebox{\textwidth}{!}{
\begin{tabular}{lcccccccccccc}
\toprule
Dataset  & In & Out & \#HL & HS & HAct & LAct & Lat & \#HHL & HHS & LRL & LR & Opt \\
\midrule
Shenzhen   & 2 & 2 & 6 & 256 & \texttt{Sine} & \texttt{Softmax} & 64 & 1 & 64 & $10^{-2}$ & $5\times10^{-6}$ & \texttt{AdamW} \\
OASIS-4    & 2 & 5 & 10 & 256 & \texttt{Sine} & \texttt{Softmax} & 2048 & 1 & 64 & $10^{-2}$ & $5\times10^{-6}$ & \texttt{AdamW}  \\
SkullBreak & 3 & 2 & 6 & 256 & \texttt{Sine} & \texttt{Softmax} & 64 & 1 & 64 & $10^{-2}$ & $5\times10^{-6}$ & \texttt{AdamW}  \\
fastMRI    & 2 & 1 & 10 & 256 & \texttt{Sine} & \texttt{Sigmoid} & 4096 & 1 & 64 & $10^{-2}$ & $5\times10^{-6}$ & \texttt{AdamW}  \\
Ultrasound & 2 & 1 & 10 & 256 & \texttt{Sine} & \texttt{Sigmoid} & 64 & 1 & 64 & $10^{-2}$ & $5\times10^{-6}$ & \texttt{AdamW} \\
\bottomrule
\end{tabular}
}
\end{table*}

% ------------------------------------------------
\begin{table}[h]
\centering
\caption{Training configuration for the neural operator (NO) mapping input to output INRs.}
\label{tab:neural_operator}
\begin{tabular}{lcccccccccc}
\toprule
Dataset & In & Out & \#HL & HD & Act & Res & LR & Opt & WD & Dropout \\
\midrule
Shenzhen   & 64 & 64 & 1 & 128 & \texttt{SiLU} & \texttt{True} & $10^{-4}$ & \texttt{AdamW} & $10^{-4}$ & 0 \\
OASIS-4    & 2048 & 2048 & 3 & 2048 & \texttt{SiLU} & \texttt{True} & $10^{-4}$ & \texttt{AdamW} & $10^{-4}$ & 0.2 \\
SkullBreak & 64 & 64 & 1 & 2048 & \texttt{SiLU} & \texttt{True} & $10^{-4}$ & \texttt{AdamW} & $10^{-4}$ & 0 \\
fastMRI    & 4096 & 4096 & 1 & 2048 & \texttt{SiLU} & \texttt{True} & $5\times10^{-5}$ & \texttt{AdamW} & $10^{-4}$ & 0.1 \\
Ultrasound & 64 & 64 & 3 & 512 & \texttt{ReLU} & \texttt{True} & $7\times10^{-4}$ &  \texttt{AdamW} & $10^{-5}$ & 0 \\
\bottomrule
\end{tabular}
\end{table}

\begin{table}[!h]
\centering
\caption{Dataset information including training, validation, and test splits, image sizes, and number of points/slices used for each experiment.}
\label{tab:datasets}
\begin{tabular}{lccccc}
\toprule
Dataset & Image Size & Train & Val & Test & Num. Points \\
\midrule
Shenzhen \cite{Jaeger2014-kh} & $200 \times 200$ & 0.8 & 0.1 & 0.1 & 40000 \\
OASIS-4 \cite{Marcus2007-or} & $160 \times 192$ & 0.7 & 0.15 & 0.15 & 30720 \\
SkullBreak \cite{Kodym2021-xv} & $512 \times 512 \times 512$ & From \cite{Kodym2021-xv} & From \cite{Kodym2021-xv} & From \cite{Kodym2021-xv} & 40000 \\
fastMRI \cite{Knoll2020-iu} & $ 128 \times 128$ & 0.8 & 0.1 & 0.1 & 10000 \\
Ultrasound (in-house) & $64 \times 64$ & 0.8 & 0.1 & 0.1 & 1820 \\
\bottomrule
\end{tabular}
\end{table}

\section*{Learning Implicit Representations}

While the main paper focuses on the performance of \textsc{NOIR} on each dataset and its corresponding task, an important component of the framework is ensuring that both the input and output INRs accurately learn the underlying signals. In particular, the latent representations optimized for each new sample should enable the INR to provide a faithful functional approximation of the data. To verify this property, we reconstruct the signals directly from the trained INRs and compare the reconstructed images with the corresponding ground truth samples. Table~\ref{tab:inr_reconstruction} reports the reconstruction quality across datasets in terms of PSNR and SSIM for both the input and output INRs, highlighting that the learned representations preserve the relevant information required to accurately model the signals.

\begin{table}[!h]
\centering
\caption{Reconstruction performance of the input and output INRs across datasets.}
\label{tab:inr_reconstruction}
\begin{tabular}{lcccccccc}
\toprule
 & \multicolumn{4}{c}{Input INR} & \multicolumn{4}{c}{Output INR} \\
\cmidrule(lr){2-4} \cmidrule(lr){5-9}
Dataset & PSNR $\uparrow$ & SSIM $\uparrow$ & DSC & IoU & PSNR $\uparrow$ & SSIM $\uparrow$ & DSC $\uparrow$ & IoU $\uparrow$\\
\midrule
Shenzhen \cite{Jaeger2014-kh}  & 22.72  & 0.66 & -- & -- & -- & -- & 0.98 & 0.97 \\
OASIS-4 \cite{Marcus2007-or}   & 24.76 & 0.79 & -- & -- & -- & --  & 0.91 & 0.84 \\
SkullBreak \cite{Kodym2021-xv} & -- & -- & 0.87 & 0.77 & -- & -- & 0.90 & 0.83\\
fastMRI  \cite{Knoll2020-iu}  & 22.26 & 0.65 & -- & -- & 29.09 & 0.71 & -- & --\\
Ultrasound & 33.77 & 0.94 & -- & -- & 33.48 & 0.70 &  --  & --\\
\bottomrule
\end{tabular}
\end{table}

\begin{figure}[!t]
\centering
\includegraphics[width=0.8\linewidth]{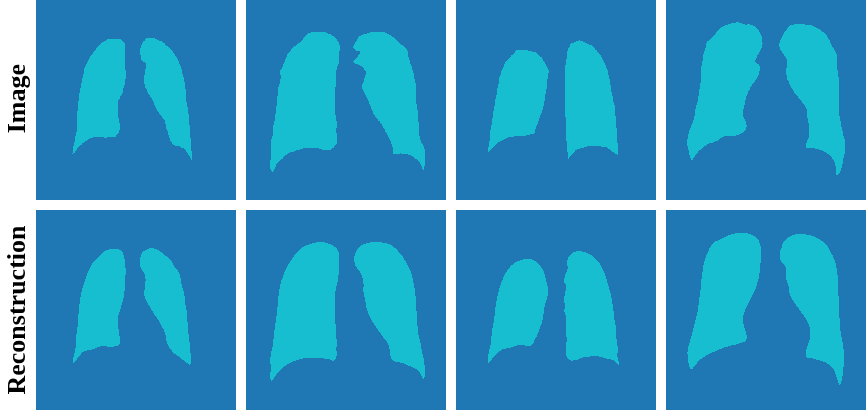}
\caption{Reconstruction results from the output INR for the Shenzhen dataset. Each example shows the ground truth image and the reconstruction produced by the output INR.}
\label{fig:recon_shenzhen}
\end{figure}

\begin{figure}[!t]
\centering
\includegraphics[width=0.8\linewidth]{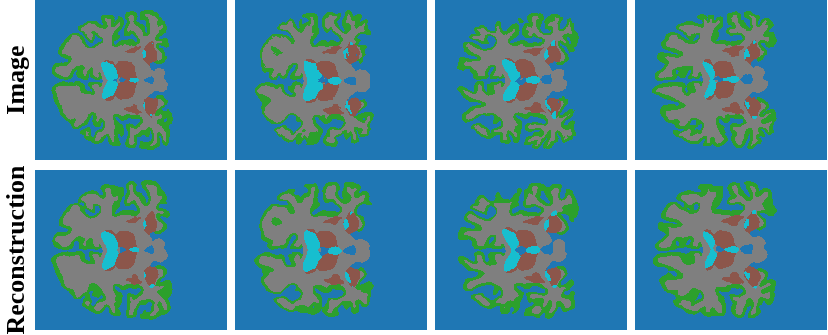}
\caption{Reconstruction results from the output INR for the OASIS-4 dataset. Ground truth segmentations are compared with the corresponding INR reconstructions.}
\label{fig:recon_oasis}
\end{figure}

\begin{figure}[!t]
\centering
\includegraphics[width=0.8\linewidth]{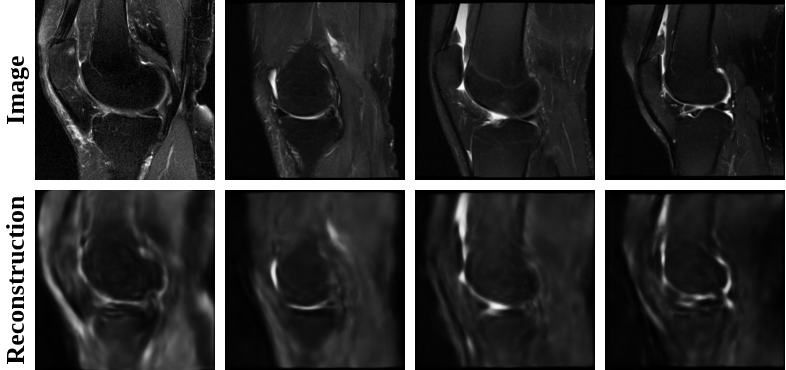}
\caption{Reconstruction results from the output INR for the fastMRI dataset. The reconstructed images are compared with the ground truth MRI slices.}
\label{fig:recon_fastmri}
\end{figure}

\begin{figure}[!t]
\centering
\includegraphics[width=0.8\linewidth]{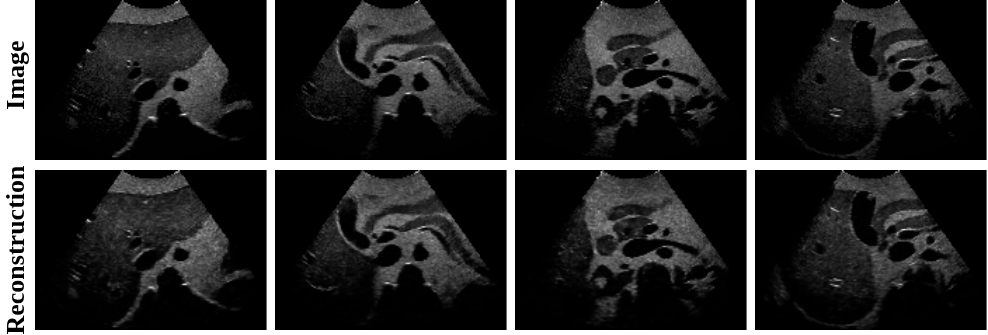}
\caption{Reconstruction results from the output INR for the ultrasound dataset, showing the ability of the INR to approximate the ultrasound signal from the learned latent representation.}
\label{fig:recon_ultrasound}
\end{figure}

\noindent In the following figures, we present qualitative reconstruction results obtained from the output INR, which is responsible for modeling the signals used in the downstream tasks. These visualizations allow us to assess how accurately the INR approximates the target signals by reconstructing them directly from the learned latent representations. Specifically, we show reconstruction examples for the Shenzhen dataset in Figure~\ref{fig:recon_shenzhen}, the OASIS-4 dataset in Figure~\ref{fig:recon_oasis}, the fastMRI dataset in Figure~\ref{fig:recon_fastmri}, and the Ultrasound dataset in Figure~\ref{fig:recon_ultrasound}. In each case, we compare the reconstructed signals obtained from the output INR with the corresponding ground truth samples.

% ------------------------------------------------
\section*{NOIR is an $\epsilon$-ReNO}

\begin{figure*}[t]
\centering
\begin{subfigure}{0.48\textwidth}
\centering
\includegraphics[width=\linewidth]{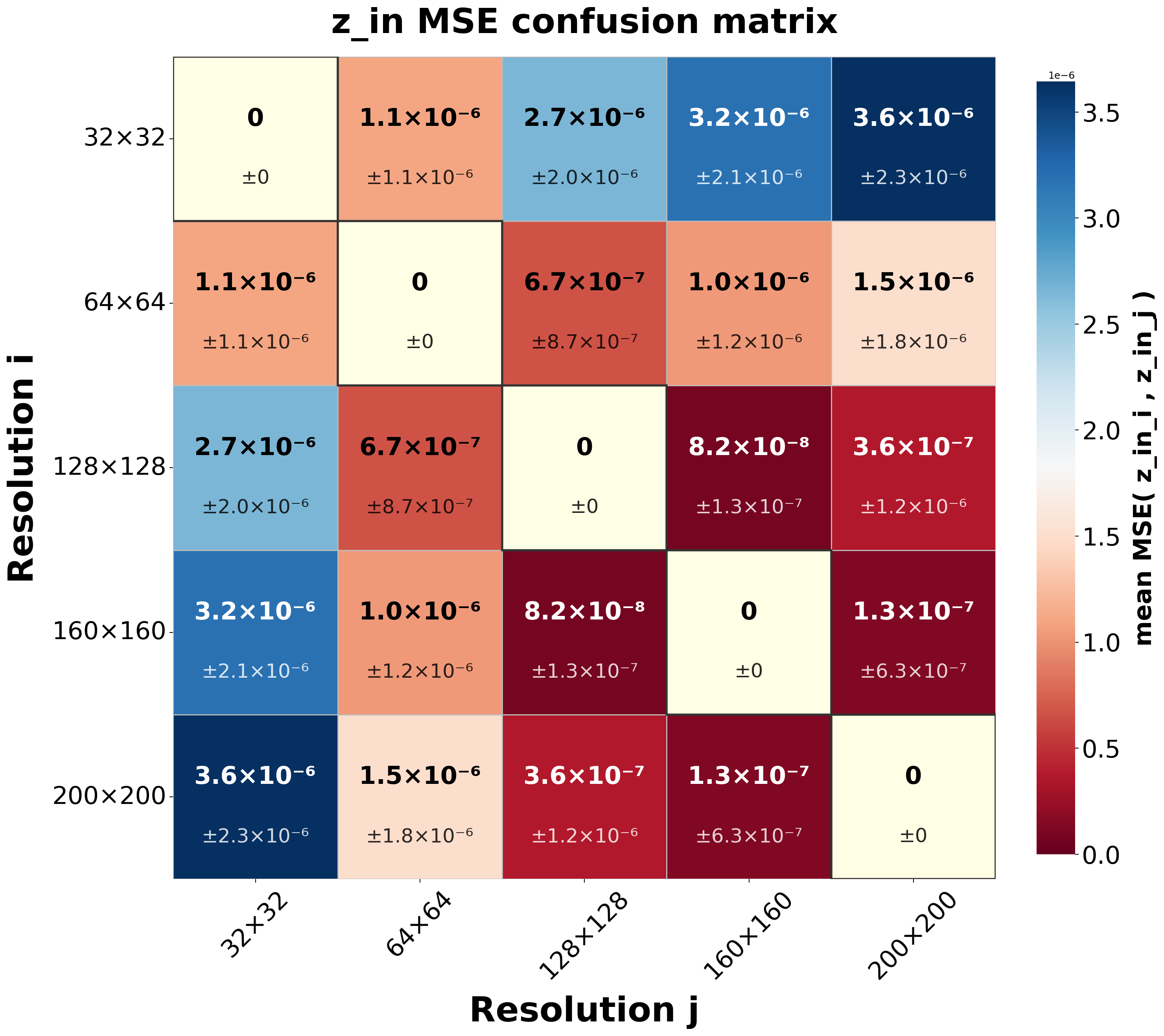}
\caption{Pairwise MSE between optimized input modulation vectors $z_{\text{in}}^{(r)}$ across resolutions.}
\label{fig:input_latent_confusion_matrix}
\end{subfigure}%
\hfill
\begin{subfigure}{0.48\textwidth}
\centering
\includegraphics[width=\linewidth]{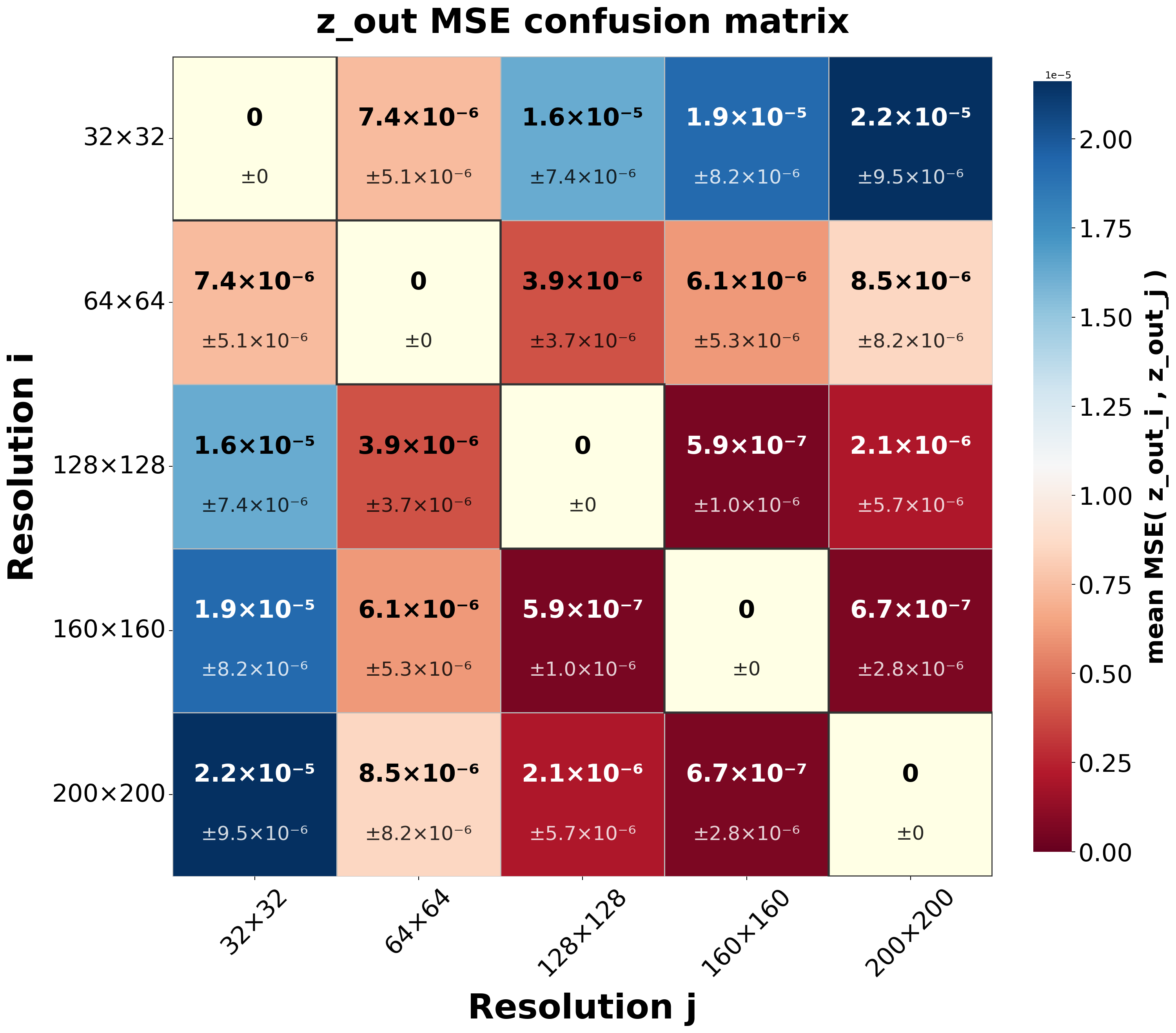}
\caption{Pairwise MSE between predicted output modulation vectors $\hat{z}_{\text{out}}^{(r)}$.}
\label{fig:output_latent_confusion_matrix}
\end{subfigure}
\caption{Confusion matrices summarizing pairwise discrepancies between latent modulation vectors obtained at resolutions $32\times32$, $64\times64$, $128\times128$, $160\times160$, and $200\times200$. Left: discrepancies between optimized input modulation vectors. Right: discrepancies between predicted output modulation vectors produced by the neural operator.}
\label{fig:latent_confusion_matrices}
\end{figure*}

To further analyze the $\epsilon$-ReNO property discussed in the paper, we provide additional quantitative visualizations of the consistency of the latent modulation vectors across multiple input discretizations. Specifically, we consider resolutions of $32\times32$, $64\times64$, $128\times128$, $160\times160$, and $200\times200$, and compute pairwise discrepancies between the modulation vectors obtained at each resolution. Figure~\ref{fig:latent_confusion_matrices} presents confusion matrices summarizing these discrepancies. Figure~\ref{fig:input_latent_confusion_matrix} shows the pairwise mean squared error (MSE) between the optimized input modulation vectors $z_{\text{in}}^{(r)}$ obtained during the inner-loop optimization at each resolution. Figure~\ref{fig:output_latent_confusion_matrix} shows the corresponding pairwise MSE between the predicted output modulation vectors $\hat{z}_{\text{out}}^{(r)}$ produced by the neural operator. These results further illustrate that both the optimized input representations and the predicted output representations remain highly consistent across different discretizations.

% ------------------------------------------------
\section*{Prediction Time}

In this supplementary material, we report the average inference time (in seconds) over the test set for each dataset for the proposed NOIR method as well as the baseline models (Table~\ref{tab:time}). For NOIR, the reported time corresponds to the sum of the time required to optimize the input INR (20 iterations in the inner loop) and the forward prediction time of the Neural Operator (NO). The resulting NO prediction is then used to modulate the output INR, enabling reconstruction of the image at arbitrary resolution. Alternatively, the NO prediction can be used directly as a modality for downstream tasks. In all cases, NOIR's prediction time remains below one second, except for the fastMRI dataset. This increase is due to the larger latent representation used for this dataset (4096). Nevertheless, NOIR remains significantly faster than the DDPM baseline, which requires approximately five seconds for the translation task.

\begin{table}[h]
\centering
\caption{Prediction time comparison (in seconds).}
\label{tab:time}
\begin{tabular}{lcccccc}
\toprule
Dataset & NOIR & U-Net & ViT & AttU-Net & FNO & DDPM \\
\midrule
Shenzhen  \cite{Jaeger2014-kh} & $8.30\times10^{-1}$ & $3.06\times10^{-3}$ & $3.07\times10^{-3}$ & $4.72\times10^{-3}$ & $2.94\times10^{-3}$ & --  \\
OASIS-4  \cite{Marcus2007-or}  & $2.41\times10^{-1}$ & $2.69\times10^{-3}$ & $5.75\times10^{-3}$ & $3.57\times10^{-3}$ & $2.63\times10^{-3}$ & -- \\
SkullBreak \cite{Kodym2021-xv} & $3.45\times10^{-1}$ & $2.13\times10^{-1}$ & $1.06\times10^{-1}$ & $2.38\times10^{-1}$ & $1.23\times10^{-1}$ & --  \\
fastMRI \cite{Knoll2020-iu}   & $1.58$ & $1.89\times10^{-3}$ & $2.71\times10^{-3}$ & $2.78\times10^{-3}$ & $1.59\times10^{-3}$ & $5.27$ \\
Ultrasound & $5.93\times10^{-1}$ & $3.00\times10^{-3}$ & $4.04\times10^{-3}$ & $4.69\times10^{-3}$ & $1.62\times10^{-3}$ & $2.72\times10^{-1}$ \\
\bottomrule
\end{tabular}
\end{table}

% ------------------------------------------------
\section*{Baseline Methods}

We provide additional implementation details for the baseline models used in our experiments. All models are implemented in both two-dimensional and three-dimensional variants depending on the dimensionality of the input signals. Inputs are resized to a fixed spatial resolution and normalized to zero mean and unit variance before being provided to the networks.

The training objective depends on the type of output signal. For segmentation and shape completion tasks, models are trained using a combination of cross-entropy loss and Dice loss with equal weighting. For image synthesis and image translation tasks, models are trained using a mean squared error (MSE) loss between the predicted output and the ground truth signal. In all experiments, optimization is performed using the \texttt{AdamW} optimizer with a learning rate of $10^{-4}$ and weight decay of $10^{-4}$.

\paragraph{U-Net.}
As a convolutional baseline, we implemented a standard U-Net architecture \cite{Ronneberger2015-ql} following the classical encoder–decoder design with skip connections between corresponding resolution levels. Each encoder block consists of two convolutional layers with kernel size $3\times3$ (or $3\times3\times3$ in the 3D case), followed by batch normalization and \texttt{ReLU} activations. Downsampling is performed using max-pooling layers. The decoder mirrors the encoder using transposed convolutions (or trilinear upsampling in the 3D case) followed by convolutional blocks. Skip connections concatenate encoder and decoder feature maps at matching resolutions. The network starts with $32$ feature channels which are doubled at each downsampling stage, resulting in feature sizes of $32, 64, 128, 256,$ and $512$ in the deepest layer. In the 3D version used for volumetric reconstruction tasks, the architecture follows the same structure but uses volumetric convolutions and trilinear upsampling, with a base feature size of $16$ channels to control memory usage. A final $1\times1$ (or $1\times1\times1$) convolution produces the output prediction.

\paragraph{Vision Transformer (ViT).}
As a transformer-based baseline, we implement a Vision Transformer architecture \cite{Dosovitskiy2020-ig}. The input image or volume is first divided into non-overlapping patches using a strided convolutional embedding layer. Each patch is projected to a token embedding space and augmented with learnable positional encodings. The sequence of tokens is then processed by a stack of transformer encoder blocks composed of multi-head self-attention layers and feed-forward multilayer perceptrons with \texttt{GELU} activations, using a pre-layer normalization configuration. In the 2D case, the model uses an embedding dimension of $192$, $8$ transformer layers, and $3$ attention heads. In the 3D variant, volumetric patches are extracted and processed using the same transformer architecture. To obtain dense predictions, the token sequence is reshaped into a spatial feature grid and passed through a hierarchical convolutional decoder with progressive upsampling stages. Skip projections from intermediate transformer layers are used to guide the decoder. The decoder uses feature dimensions of $128, 64, 32,$ and $16$ channels before a final convolutional layer produces the output signal.

\paragraph{Attention U-Net.}
We implemented an Attention U-Net architecture that extends the standard U-Net by introducing attention gates in the skip connections between encoder and decoder stages \cite{Oktay2018-hr}. The encoder and decoder structures follow the same convolutional design as the U-Net baseline. At each resolution level, the feature maps from the encoder are passed through an additive attention mechanism before being concatenated with the decoder features. These attention gates learn spatial attention coefficients that highlight task-relevant regions and suppress irrelevant activations, allowing the network to focus on informative structures in the input. The model uses the same feature scaling strategy as the U-Net baseline, starting from $32$ channels in the 2D case and $16$ channels in the 3D case, doubling the number of channels at each downsampling stage. In the volumetric implementation, all convolutional operations are replaced with their three-dimensional counterparts.

\paragraph{Fourier Neural Operator (FNO).}
As an operator-learning baseline, we implement a Fourier Neural Operator architecture \cite{Li2020-ss}. The model follows a lifting–operator–projection design. First, the input signal is lifted to a higher-dimensional latent representation using a $1\times1$ (or $1\times1\times1$) convolution. The lifted representation is then processed by a stack of Fourier operator blocks that combine spectral and spatial transformations. In each block, the feature maps are transformed to the Fourier domain, where a truncated set of low-frequency modes is retained. A learnable complex weight tensor is applied to these modes before transforming the representation back to the spatial domain using the inverse Fourier transform. In parallel, a residual pointwise convolution operates in the spatial domain to capture local interactions. The outputs of the spectral and spatial paths are combined and passed through a nonlinear activation. In the 2D configuration, we retain $16\times16$ Fourier modes and use four operator layers with hidden feature dimension $64$. In the 3D variant, volumetric Fourier transforms are applied and $12\times12\times12$ spectral modes are retained to maintain computational efficiency. The final representation is projected to the desired output space using a small convolutional head.

\paragraph{Denoising Diffusion Probabilistic Model (DDPM).}
We implement a conditional Denoising Diffusion Probabilistic Model following Ho et al. \cite{ho2020denoising} for image-to-image translation. The model learns to iteratively denoise a Gaussian-corrupted target image, conditioned on a source image. During training, the target image is progressively noised according to a forward diffusion process defined by a linear variance schedule over $T=1000$ timesteps. At each training step, a random timestep $t$ is uniformly sampled, the target is noised to that level, and the noised target is channel-wise concatenated with the conditioning image to form a two-channel input. The model is trained to predict the added noise (epsilon parameterization) using a mean squared error loss. The denoising network is a U-Net architecture with sinusoidal timestep embeddings injected via Feature-wise Linear Modulation like scale-and-shift conditioning in each residual block. The encoder–decoder structure uses a channel multiplier schedule of $(1,2,4,8)$ starting from 64 base channels, with two residual blocks per resolution level. Multi-head self-attention with four heads is applied at resolutions $32\times32,16\times16$, and $8\times8$, and additionally at the bottleneck. Down- and upsampling are performed using residual blocks with learned strided or transposed convolutions. The encoder and decoder stages are connected via concatenative skip connections. At inference, sampling starts from pure Gaussian noise and iteratively denoises over 1000 steps using the learned reverse process, concatenating the conditioning image at each step.

\end{document}